\PassOptionsToPackage{table}{xcolor}
\documentclass{article}
\usepackage{iclr2027_conference,times}

\usepackage{amsmath,amsfonts,bm}

\def\Figref#1{Figure~\ref{#1}}

\def\eqref#1{equation~\ref{#1}}

\def\1{\bm{1}}

\DeclareMathAlphabet{\mathsfit}{\encodingdefault}{\sfdefault}{m}{sl}
\SetMathAlphabet{\mathsfit}{bold}{\encodingdefault}{\sfdefault}{bx}{n}

\usepackage{hyperref}
\usepackage{url}

\usepackage[disable,textsize=tiny]{todonotes}
\usepackage[table]{xcolor}
\usepackage{booktabs}
\usepackage{multirow}
\usepackage{makecell}
\usepackage{siunitx}
\usepackage{graphicx}
\usepackage{float} 
\usepackage{pifont}
\usepackage{tcolorbox}

\usepackage{tabularx}
\usepackage{wrapfig}
\usepackage{enumitem}
\usepackage{amssymb}

\definecolor{baselinecolor}{HTML}{EEEEEE}

\definecolor{lightgray}{gray}{0.9}
\definecolor{cornellred}{rgb}{0.7, 0.11, 0.11}
\definecolor{cadmiumgreen}{rgb}{0.0, 0.42, 0.24}
\definecolor{aliceblue}{rgb}{0.91, 0.94, 0.97}
\definecolor{darkblue}{rgb}{0.83, 0.89, 0.97}
\definecolor{Red7}{rgb}{0.941, 0.243, 0.243}
\definecolor{Green7}{RGB}{55, 178, 77}
\definecolor{Blue9}{rgb}{0.098,0.3,0.9}

\usepackage{pifont}

\usepackage{listings}
\usepackage[T1]{fontenc}
\usepackage{hyperref}
\hypersetup{
  linkcolor = cornellred,
  citecolor  = cadmiumgreen,
  colorlinks = true,
  urlcolor = Blue9
}

\title{Learning What Actions Change: \\ Action Effect Matching for \\ Dynamics-Faithful World Models}

\author{John Won$^1$ \quad Yonghoon Dong$^1$ \quad Jinwoo Shin$^{1,2}$\\
$^{1}$KAIST \quad $^{2}$RLWRLD\\
}

\usepackage{xcolor}
\newcommand{\method}{\textsc{AFFECT}}

\iclrfinalcopy

\begin{document}

\maketitle

\begin{abstract}
World models aim to predict environment dynamics under actions, enabling the prediction of action consequences without executing them in the real world. 
Yet despite producing increasingly realistic videos, current world models often fail to capture how small action changes affect future observations.
This limits their use in tasks such as robotic policy evaluation and model-based planning, where nearly identical actions can lead to qualitatively different outcomes.
To address this, we propose \method{}, an auxiliary training objective for learning the distribution of visual changes induced by action perturbations.
Specifically, we inject temporally correlated noise into the actions of a policy trained on task-agnostic human teleoperation. 
We then train the world model so that its generated videos respond to this noise as the recorded videos do, without requiring repeated executions from the same initial state.
Across OGBench simulation and real-world DROID and ALOHA-2 setups, \method{} makes world-model rollouts more faithful to real outcomes. 
On DROID, the Pearson correlation between policy success rates in the world model and on the real robot rises from 0.47 to 0.79, alongside up to 23\% lower FVD. 
These gains carry over to planning, where selecting actions with \method{}-trained world models improves task success by up to $8.3\%$.
\end{abstract}

\section{Introduction}
\suppressfloats[t]
\begin{figure*}[t]
\centering\small
\includegraphics[width=\linewidth]{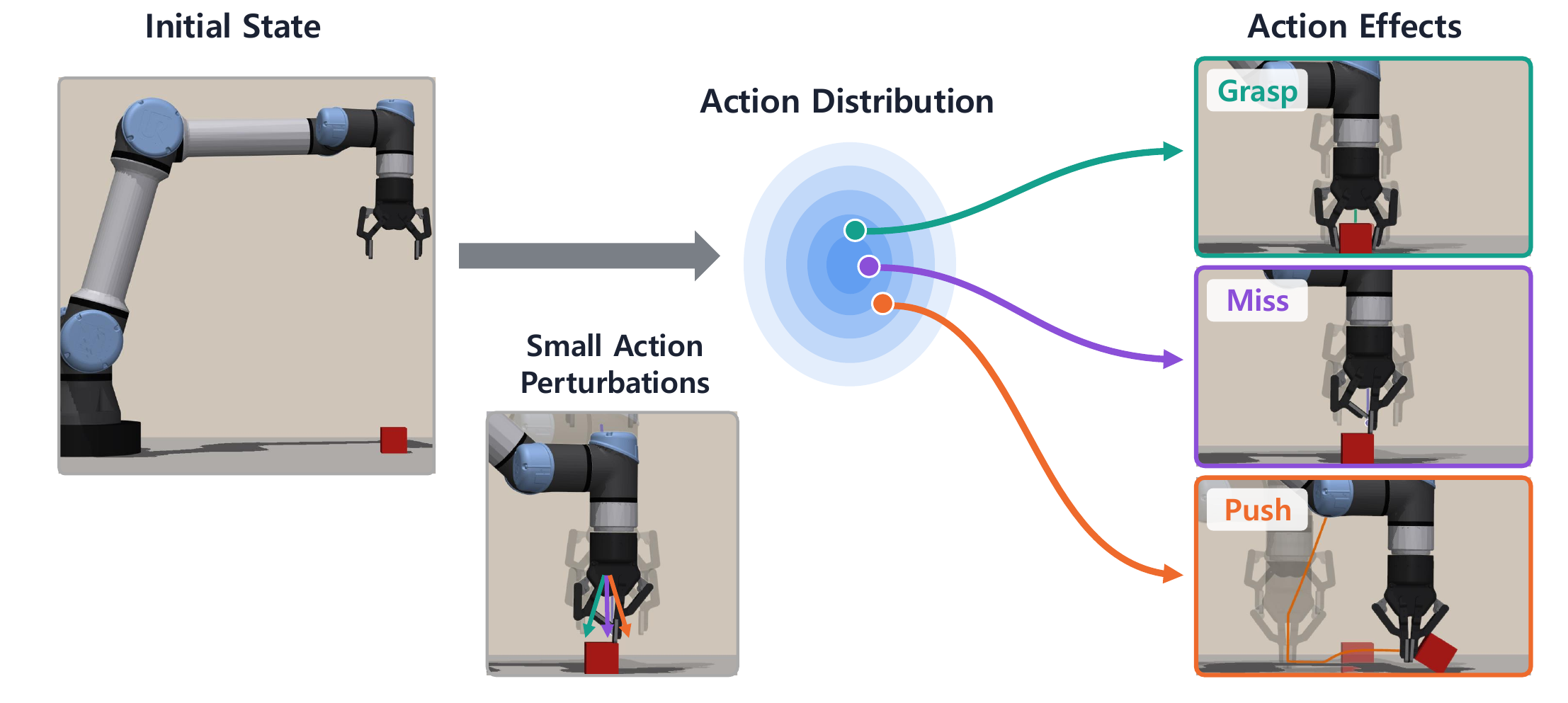}
\vspace{-15pt}
\caption{
{\bf Small action perturbations lead to distinct action effects.}
From the same initial state, actions drawn from a narrow distribution change the gripper's approach only slightly, yet produce qualitatively different outcomes, where the cube is grasped, missed, or pushed away.
A faithful world model must reproduce how these outcomes shift as the action is perturbed, rather than only rendering a plausible video for generic actions.
}
\label{fig:motivation}
\vspace{-17pt}
\end{figure*}

World models \citep{ha2018world, hafner2019planet, hafner2025dreamerv3} predict future states conditioned on action sequences without execution in the environment.
Recently, robotics world models (RWMs) \citep{gao2026dreamdojo, agarwal2026cosmos, guo2026ctrlworld} realize this by building on video diffusion foundation models \citep{wan2025wan,ali2025cosmospredict25, kong2024hunyuanvideo}, generating photorealistic, temporally coherent videos conditioned on robot actions.
This makes them a promising direction for diverse usage in evaluating policies \citep{guo2026ctrlworld, jeon2026roboworld}, planning manipulation trajectories \citep{assran2025vjepa2, gao2026dreamdojo}, or generating training data \citep{jang2025dreamgen}.

However, serving these roles requires robotics world models to faithfully reflect how small action changes alter future outcomes. For example, slight changes in reaching motion can determine whether a gripper grasps an object, misses it, or pushes it away (\Figref{fig:motivation}).
Yet existing models tend to produce visually plausible videos that ignore such differences \citep{yang2026mirabench, jiang2026robowmbench}. Since such models predict similar outcomes for actions with different real consequences, they are unreliable as policy evaluators and mislead planners into selecting the wrong actions.

Addressing this requires world models to acquire action sensitivity, and by extension, learn the relationship between action variations and resulting dynamics.
We therefore treat action sensitivity as a distributional problem.
Rather than simply reproducing recorded futures with high visual fidelity, we argue that an RWM should capture how the distribution of future observations shifts as actions vary under a given observation context. 

To learn such distributions, we introduce action variation directly into the training data. Specifically, we collect robot-environment interactions by running a policy trained on task-agnostic human teleoperation, in which an operator interacts freely with objects without a predefined task. Following \citet{lynch2019play}, we refer to such data as play data and to this policy as the play policy.

During data collection, we inject temporally correlated noise into the action chunks produced by the play policy before rollout and log the injected perturbations.
These rollouts expose how visual outcomes vary with action changes without requiring repeated executions from the same initial state or explicit outcome labels.

Building on this data, we introduce \method{}, an auxiliary training objective that teaches the world model the relationship between action changes and their visual consequences. 
\method{} explicitly uses the noise injected into the training actions to supervise how generated visual outcomes should change with those action perturbations. 
In this way, the model is directly encouraged to respond to action variation in the same way as the real environment, rather than merely generating visually plausible futures. 
Since \method{} operates on features from frozen encoders, it can be added to existing RWMs without architectural changes.

We evaluate \method{} on a simulated UR5e environment and two real-robot settings following the DROID~\citep{khazatsky2024droid} and ALOHA-2~\citep{aldaco2024aloha2} configurations, and apply it to two distinct backbones, DreamDojo~\citep{gao2026dreamdojo} and Cosmos3~\citep{agarwal2026cosmos}.
To test whether the world model reflects action differences, we measure how well its predicted policy success rates correlate with real ones.
On this measure, \method{} improves every one of the six backbone--embodiment combinations, for instance raising the correlation of DreamDojo on DROID from $0.468$ to $0.790$ and that of Cosmos3 on ALOHA-2 from $0.298$ to $0.694$.
\method{} also improves most visual metrics, including FVD reductions of up to $23\%$ on DROID and gains on every metric at both short and long horizons in simulation, suggesting that learning how actions shape outcomes also helps the model predict more realistic futures.
Finally, this improved sensitivity translates into better model-based planning, where \method{} raises task success rates by around $8\%$ on both UR5e pick-and-place and ALOHA-2 long-horizon tasks.

\section{Preliminaries}
\label{sec:preliminaries}
\textbf{World model training.}
Our aim is to develop a world model $W_\theta$ capable of visually simulating the outcomes of an arbitrary robot policy $\pi$.
We write vectors in bold lowercase, matrices and sequences in bold uppercase, and sets in calligraphic letters; Appendix~\ref{sec:icm-setting} summarizes the notation.
Formally, let $\mathcal{D}=\{(\bm{o}^{(n)}_t, \bm{a}^{(n)}_t)_{t\ge 0}\}_{n=1}^{N}$ denote a dataset of $N$ robot trajectories, each a sequence of video observations $\bm{o}_t$ and corresponding actions $\bm{a}_t\in\mathbb{R}^D$, where $t$ denotes the environment timestep and $D$ the action dimension.
The world model $W_\theta(\bm{O}_t\mid\bm{o}_t,\bm{A}_t)$ is a conditional distribution over future observations, which we train to predict future states over an $H$-frame horizon.
Specifically, it generates a future video sequence $\bm{O}_t := (\bm{o}_{t+1}, \dots, \bm{o}_{t+H})$ conditioned on the current observation $\bm{o}_t$ and action sequence $\bm{A}_t := (\bm{a}_t, \dots, \bm{a}_{t+H-1})\in\mathbb{R}^{H\times D}$.

Following recent robotics world models~\citep{gao2026dreamdojo, agarwal2026cosmos}, we build on a conditional rectified-flow video model~\citep{lipman2023flow,liu2023rectified}. We parameterize a velocity field $v_\theta(\bm{O}_t^\tau,\tau;\, \bm{o}_t,\bm{A}_t)$ and train it on $\bm{O}_t^\tau := (1-\tau)\bm{O}_t + \tau\bm{\varepsilon}$, with flow timestep $\tau \in [0,1]$ and $\bm{\varepsilon} \sim \mathcal{N}(\bm{0},\bm{I})$:
\begin{equation}
  \mathcal{L}_{\mathrm{flow}} = \mathbb{E}_{(\bm{o}_t,\bm{A}_t,\bm{O}_t)\sim\mathcal{D},\,\tau,\,\bm{\varepsilon}}\!\left[\left\| v_\theta(\bm{O}_t^\tau,\tau;\, \bm{o}_t,\bm{A}_t) - (\bm{\varepsilon} - \bm{O}_t) \right\|_2^2\right].
  \label{eq:method-flow}
\end{equation}
During inference, we initialize $\hat{\bm{O}}_t^1 \sim \mathcal{N}(\bm{0},\bm{I})$ and integrate $v_\theta$ from $\tau=1$ to $\tau=0$ with the multistep UniPC solver~\citep{zhao2023unipc} to obtain the predicted future video $\hat{\bm{O}}_t$, whose distribution defines the world model $W_\theta(\cdot \mid \bm{o}_t, \bm{A}_t)$.

\textbf{World model evaluation.}
For policy evaluation, given an initial observation $\bm{o}_0$ and a language instruction $l$, the generalist policy $\pi$ instead selects actions from the generated observations.
Starting from $\hat{\bm{o}}_0=\bm{o}_0$, the policy outputs $\bm{A}_t=\pi(\hat{\bm{o}}_t,l)$, and the world model predicts the resulting future $\hat{\bm{O}}_t$ conditioned on $\hat{\bm{o}}_t$ and $\bm{A}_t$.
The final predicted observation $\hat{\bm{o}}_{t+H}$ is then fed back into $\pi$ to select the next action sequence.
This forms a fully closed-loop system of interaction between the policy and the world model \citep{guo2026ctrlworld}, allowing long-horizon policy rollout without executing each candidate policy on a real robot.
The usefulness of these rollouts depends on whether task progress and success in the generated videos agree with those in the real environment, so that the model supports reliable comparisons between policies.

Visual quality alone is insufficient for this purpose.
A model may preserve the scene and generate plausible arm motion while missing whether the gripper makes contact with an object or whether the object moves as commanded.
Such errors can change both the predicted task outcome and the actions selected by the policy at the next step.
Evaluation therefore also needs to measure action fidelity, whether changes in the supplied actions produce the corresponding changes in robot-object interaction and task outcome.
To this end, we utilize the Pearson coefficient of success rate between real rollouts of policies and their closed-loop rollouts within world models.

\section{Method}
\label{sec:method}
A world model trained only to reconstruct recorded videos can generate realistic futures that barely depend on the supplied actions.
We introduce \method{}, an auxiliary training objective that targets this failure directly.
Rather than asking each generated video to match its recorded counterpart, \method{} asks generated videos to respond to action perturbations in the same way as recorded videos.
This requires interaction data in which actions are perturbed in a known way (Section~\ref{sec:method-data}) and a loss that measures the response to these perturbations (Section~\ref{sec:method-matching}).
Section~\ref{sec:method-statistics} then interprets this loss statistically.

\subsection{Data generation via perturbation play}
\label{sec:method-data}

Play data \citep{lynch2019play, yin2026playworld}, in which the robot interacts freely with its environment without task instructions, covers a broad range of interactions.
We train a play policy on human-teleoperated play trajectories and roll it out autonomously, which requires no scene resets or task-specific procedures and therefore scales to large amounts of data.
To vary actions around what the policy would do, we inject temporally correlated noise into each action chunk and execute $\bm{A}_t = \mathrm{clip}(\bar{\bm{A}}_t + \bm{U}_t)$, where $\bar{\bm{A}}_t\in\mathbb{R}^{H\times D}$ is the nominal chunk from the play policy and $\bm{U}_t=(\bm{u}_t,\dots,\bm{u}_{t+H-1})\in\mathbb{R}^{H\times D}$ is the injected noise.
We refer to each chunk, together with its context $\bm{o}_t$ and recorded future $\bm{O}_t$, as a \emph{window}.

For each window, we log the perturbation $\bm{U}_t$, its noise-distribution parameters, and the nominal and executed commands.
We log $\bm{U}_t$ before clipping so that the logged samples follow the specified noise distribution.
The world model receives only $\bm{o}_t$ and $\bm{A}_t$ as inputs, and $\bm{U}_t$ is used only in the \method{} loss.
Because this loss compares recorded and generated videos across rollouts, each window needs a single recorded future, and no repeated executions from the same starting state are required.
Appendix~\ref{sec:icm-data} details the collection protocol.

\begin{figure*}[t]
\centering\small
\includegraphics[width=\linewidth]{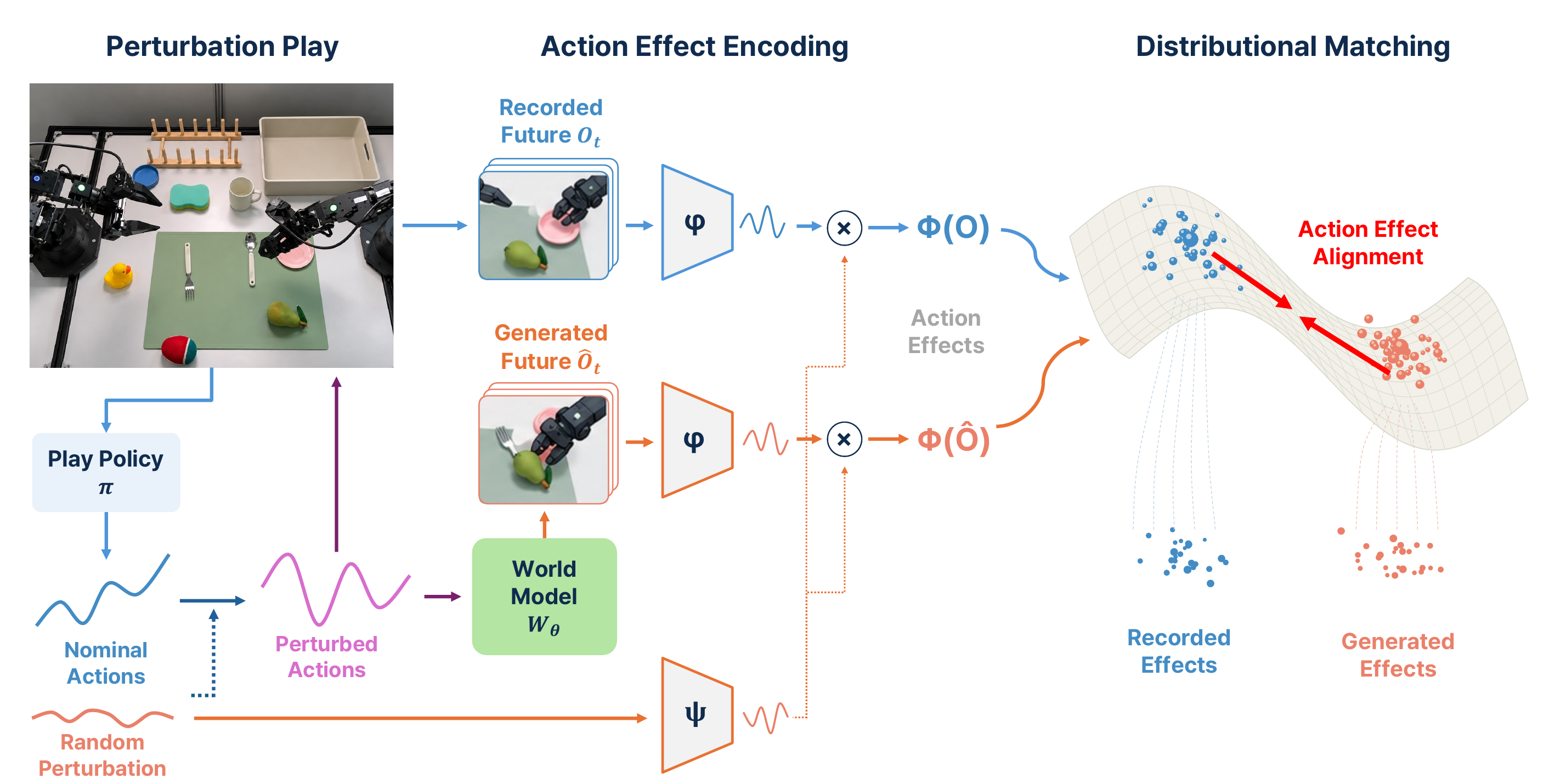}
\vspace{-10pt}
\caption{
{\bf Overview of \method{}.}
Frozen encoders $\phi,\psi$ extract video features from both the recorded and generated futures, and perturbation features from the logged action perturbations.
Pairing the centered perturbations with each set of video features gives the recorded and predicted action effects ($\Phi(\bm{O}), \Phi(\hat{\bm{O}})$), which measure how visual change moves with the perturbation across rollouts.
The \method{} loss optimizes over the distance between the two distributions, so that generated futures respond to perturbations in the same direction and with the same strength as real ones.
}
\label{fig:diff_steps}
\vspace{-10pt}
\end{figure*}

\subsection{Matching action effects with frozen encoders}
\label{sec:method-matching}
The data of Section~\ref{sec:method-data} pair each recorded future with the perturbation that produced it.
Across many windows, these pairs reveal how the scene responds to actions.
If stronger pushes tend to move an object farther, for instance, then recorded windows with a stronger push will on average show the object farther away.
We want generated videos to show the same tendency.
To measure it, we summarize the perturbation and the video by a few features each and ask how the two sets of features vary together.

\textbf{Frozen feature encoders.}
We compute these features with two fixed encoders, $\psi(\bm{U}_t)\in\mathbb{R}^{K_\psi}$ for the perturbation and $\phi(\bm{O}_t)\in\mathbb{R}^{K_\phi}$ for the video.
Each output is a simple, hand-specified quantity.
A perturbation feature might be the signed or squared sum of the noise over the window, and a video feature might be a change in a per-channel mean of the video latent or a projection of the latent change.
Appendix~\ref{sec:icm-bank} lists the full feature bank.
We keep both encoders frozen throughout training.
If $\phi$ were trained jointly with the world model, it could drive the loss to zero by outputting a constant, and the model would learn nothing about actions.

\textbf{Centering the perturbation features.}
Before relating the two sets of features, we need the perturbation features to vary around zero, and some of them do not.
The squared noise, for example, is always positive.
Multiplying it with a video feature would then reward matching that feature even when the perturbation has no effect on it, so the loss would end up supervising static content such as the background.
We avoid this by subtracting the expected value of each perturbation feature,
\begin{equation}
\bm{g}_t := \psi(\bm{U}_t)-\mathbb{E}[\psi(\bm{U}_t)\mid\mathcal{H}_t]\in\mathbb{R}^{K_\psi},
\label{eq:method-centering}
\end{equation}
where $\mathcal{H}_t$ is the information available before the window, namely $\bm{o}_t$ and the declared parameters of the perturbation distribution.
Since the noise is drawn from these declared parameters, this expected value can be computed directly from them.
After centering, each entry of $\bm{g}_t$ is positive when the corresponding perturbation feature is larger than expected and negative when it is smaller.

\textbf{Matching action effects.}
With centered perturbation features in hand, we can measure how a video feature responds to the perturbation.
For perturbation feature $i$ and video feature $j$, we define the \emph{action effect} of the recorded videos as $\mathbb{E}[g_{t,i}\,\phi_j(\bm{O}_t)]$, the average over training windows of the centered perturbation feature times the video feature.
It is positive when larger-than-expected perturbations tend to come with larger values of the video feature, and negative when they tend to come with smaller ones.
Replacing $\bm{O}_t$ with a generated future $\hat{\bm{O}}_t$ gives the corresponding action effect of the world model, which also averages over the model's samples.
\method{} trains the world model so that the two agree for every pair of features,
\begin{equation}
\mathcal{L}_{\mathrm{AFFECT}} := \sum_{i=1}^{K_\psi}\sum_{j=1}^{K_\phi}\left(\frac{\mathbb{E}\bigl[g_{t,i}\bigl(\phi_j(\bm{O}_t)-\phi_j(\hat{\bm{O}}_t)\bigr)\bigr]}{s_{ij}}\right)^{2},
\qquad
\mathcal{L}:=\mathcal{L}_{\mathrm{flow}}+\lambda\mathcal{L}_{\mathrm{AFFECT}}.
\label{eq:method-objective}
\end{equation}
Each scale $s_{ij}>0$ is the standard deviation of $g_{t,i}\,\phi_j(\bm{O}_t)$ on recorded windows, fixed before \method{} training, so that pairs with large raw values do not dominate the sum.
The weight $\lambda>0$ balances the auxiliary loss against the flow loss.

\textbf{Computing the loss during training.}
The loss needs generated futures, and how we produce them matters.
The rectified-flow model already predicts the clean video in one step from a noised version of $\bm{O}_t$, but this prediction is built from the recorded future itself and would match it whether or not the model uses the action.
We instead generate $\hat{\bm{O}}_t$ from fresh Gaussian noise with a short differentiable Euler sampler, conditioning on the same $\bm{o}_t$ and $\bm{A}_t$ as the recorded window but never on $\bm{O}_t$.
Because this sampling is expensive, we apply the loss only every few iterations.
Finally, the loss squares an average over windows, and squaring a minibatch average in its place would also penalize how much the generated futures vary, which pushes the model toward mode collapse.
We therefore compute $\mathcal{L}_{\mathrm{AFFECT}}$ on minibatches in a form that removes this extra term (Appendix~\ref{sec:icm-estimator}).

\subsection{Statistical interpretation of \method{}}
\label{sec:method-statistics}

Section~\ref{sec:method-matching} built \method{} from two operations, centering the perturbation features and multiplying them with video features.
Both have standard statistical readings, which we now give to make precise what the loss measures and what it leaves unconstrained.

\textbf{Action effects as covariances.}
Centering is what turns an action effect into a covariance.
Because $\bm{g}_t$ has zero mean given $\mathcal{H}_t$, the action effect equals the conditional covariance between the perturbation feature and the video feature, averaged over contexts,
\begin{equation}
  \mathbb{E}[g_{t,i}\,\phi_j(\bm{O}_t)]
  = \mathbb{E}\!\left[\operatorname{Cov}\!\left(\psi_i(\bm{U}_t),\phi_j(\bm{O}_t)
      \mid\mathcal{H}_t\right)\right].
  \label{eq:method-covariance}
\end{equation}
An action effect therefore measures how strongly video feature $j$ moves with perturbation feature $i$ within the same scene, rather than across different scenes.
This view also makes the role of centering precise.
Any video feature whose value is fixed by the context alone has zero covariance with the perturbation, so errors that depend only on the scene, such as rendering a scene slightly too dark, cancel between recorded and generated videos (Appendix~\ref{sec:icm-analysis-centering}).
Each term of Eq.~\ref{eq:method-objective} thus compares the direction and strength of the response to actions while largely ignoring what the scene looks like.

A simple linear model shows what matching these covariances enforces.
Suppose a video feature responds to a scalar perturbation with coefficient $\beta$ in the recorded videos and $\hat\beta$ in the generated ones.
The corresponding term of the loss is then proportional to $(\beta-\hat\beta)^2$, so it vanishes exactly when the model reproduces the recorded response, and it grows the more the model ignores the action, overreacts, or responds in the wrong direction (Appendix~\ref{sec:icm-rationale-response}).
More generally, a model that reproduces the recorded distribution of futures attains zero loss whenever $\bm{o}_t$ and $\bm{A}_t$ carry all the information about the outcome that $\bm{U}_t$ and $\mathcal{H}_t$ carry.
Our data satisfy this only approximately, because the play policy observes more of the scene than the world model does (Appendix~\ref{sec:icm-setting}).

\begin{table*}[t]
\centering
\caption{{\bf Simulation evaluation results}
Video fidelity metrics (PSNR, LPIPS, FVD) for single-window short horizon video generation and autoregressive long horizon video generation in the simulated UR5e setting. We also report the Pearson correlation of task success rates on pick-and-place tasks in OGBench between real and generated rollouts.}
\vspace{0.10in}
\newcolumntype{a}{>{\columncolor{green! 10}}c}
\begin{tabular}{l a ccc ccc}
\toprule
& \multicolumn{1}{c}{} & \multicolumn{3}{c}{Short Horizon} & \multicolumn{3}{c}{Long Horizon} \\
\cmidrule(lr){3-5} \cmidrule(lr){6-8}
Method & Pearson $\uparrow$ & 
PSNR $\uparrow$ & LPIPS $\downarrow$ & FVD $\downarrow$ & 
PSNR $\uparrow$ & LPIPS $\downarrow$ & FVD $\downarrow$ \\
\midrule
iVideoGPT & 0.334 & 
21.53 & 0.49 & 1134 &
20.20 & 0.53 & 1314 \\
CoCo & 0.130 & 
21.69 & 0.49 & 1094 &
20.29 & 0.52 & 1331 \\
Ctrl-World & 0.310 & 
24.99 & 0.10 & 156 &
24.27 & 0.11 & 196 \\
\midrule
Cosmos3 & 0.091 & 
19.07 & 0.19 & 470 &
18.72 & 0.22 & 562 \\
{\bf + Ours} & 0.334 & 
19.95 & 0.18 & 463 &
19.21 & 0.20 & 540 \\
\midrule
DreamDojo & 0.451 & 
26.03 & 0.06 & 85 &
24.87 & 0.08 & 108 \\
{\bf + Ours} & {\bf 0.686} & 
{\bf 26.78} & {\bf 0.05} & {\bf 73} &
{\bf 25.58} & {\bf 0.07} & {\bf 93 } \\
\bottomrule
\end{tabular}
\label{Tab:tab_sim_ur5e}
\end{table*}

\textbf{Feature matching on action effects.}
Taken together, the terms of the loss relate \method{} to feature matching.
Standard feature matching compares the average video features of recorded and generated futures.
\method{} instead compares how these averages change with the perturbation, and the difference matters in practice.
Consider a model that moves an object left whenever the real object moves right.
The perturbations and the object positions then have the same distributions in both cases, so matching average features cannot detect the error, while the product with the perturbation flips its sign and exposes it.
Formally, we collect the scaled products into a feature map $\Phi_t(\bm{O})\in\mathbb{R}^{K_\psi K_\phi}$ with entries $g_{t,i}\,\phi_j(\bm{O})/s_{ij}$ and rewrite the loss as
\begin{equation}
  \mathcal{L}_{\mathrm{AFFECT}}
  = \left\|\mathbb{E}[\Phi_t(\bm{O}_t)]-\mathbb{E}[\Phi_t(\hat{\bm{O}}_t)]\right\|_2^2.
  \label{eq:method-mmd}
\end{equation}
This is a finite-feature maximum mean discrepancy (MMD)~\citep{gretton2012mmd}, the objective behind moment-matching generative models~\citep{li2015gmmn,dziugaite2015mmdopt,ren2016cgmmn}.
Two consequences follow from this view.
First, the minibatch form of Section~\ref{sec:method-matching} is the paired form of the standard unbiased estimator of the squared MMD, which avoids the variance term of a naive plug-in estimate.
Second, a small loss bounds not only each individual term but also every normalized linear combination of the terms (Appendix~\ref{sec:icm-analysis-products}).

\section{Experiments}
\label{sec:experiments}
We validate \method{} on simulated and real-robot perturbation play. We design our experiments to investigate the following questions:
\begin{itemize}[leftmargin=*,itemsep=0pt,topsep=2pt]
  \item Does \method{} improve the visual/action fidelity of generated futures? (Tables~\ref{Tab:tab_sim_ur5e} and~\ref{Tab:tab_real})
  \item Does a more action-faithful world model help select better actions at test time? (\Figref{fig:planning})
  \item Do the frozen features actually see the action effect? (Figure~\ref{fig:effect-signal})
\end{itemize}

\begin{table}[t]
\centering\small
\caption{{\bf Real-world robot evaluation results}
Video fidelity metrics for autoregressive long horizon video generation in the domain of real-world DROID and ALOHA-2 hardware configurations. We also report the Pearson correlation of task success rates between real and generated rollouts.
}
\vspace{0.10in}
\newcolumntype{a}{>{\columncolor{green! 10}}c}
\begin{tabular}{l ccca c ccca}
\toprule
& \multicolumn{4}{c}{DROID (Real)} &
& \multicolumn{4}{c}{ALOHA-2 (Real)} \\
\cmidrule(lr){2-5}
\cmidrule(lr){7-10}
Method & 
PSNR $\uparrow$ & LPIPS $\downarrow$ & FVD $\downarrow$ & Pearson $\uparrow$ &
& PSNR $\uparrow$ & LPIPS $\downarrow$ & FVD $\downarrow$ & Pearson $\uparrow$ \\
\midrule
iVideoGPT &
16.16 & 0.52 & 1946 & 0.151 &&
16.77 & 0.52 & 3808 & 0.155 \\
CoCo &
18.13 & 0.50 & 1676 & 0.212 &&
19.09 & 0.49 & 3959 & 0.225 \\
Ctrl-World &
17.48 & 0.36 & 1141 & 0.236 &&
20.32 & 0.11 & 396 & 0.399 \\
\midrule
Cosmos3 &
18.69 & 0.24 & 910 & 0.425 &&
22.94 & 0.08 & 333 & 0.298 \\
{\bf + Ours} &
{\bf 19.67} & 0.22 & 744 & 0.583 &&
24.05 & {\bf 0.05} & 257 & {\bf 0.694} \\
\midrule
DreamDojo &
18.53 & 0.16 & 807 & 0.468 &&
24.20 & 0.06 & {\bf 183} & 0.325 \\
{\bf + Ours} &
19.04 & {\bf 0.14} & {\bf 622} & {\bf 0.790} &&
{\bf 24.91} & {\bf 0.05} & 188 & 0.490 \\
\bottomrule
\end{tabular}
\label{Tab:tab_real}
\end{table}

\subsection{Robotics world models for policy evaluation}
\label{sec:exp-setup}

\textbf{Embodiments and Datasets.}
We consider three embodiments, each collected with the perturbation play protocol of Section~\ref{sec:method-data}. 
In simulation, we use a UR5e manipulation environment from OGBench~\citep{park2025ogbench}. 
On real robots, we use a Franka Research 3 in the hardware configuration proposed by DROID~\citep{khazatsky2024droid} and a bimanual ALOHA-2~\citep{aldaco2024aloha2} setup.  
Every episode is a single unpaired rollout, since \method{} never uses repeated executions from the same start. 
Further details are in Appendix~\ref{sec:icm-data}.

\textbf{Implementation details.}
We apply \method{} to two action-conditioned rectified-flow backbones, DreamDojo~\citep{gao2026dreamdojo} and Cosmos3~\citep{agarwal2026cosmos}, which represent the state-of-the-art among robotics world models. 
Both are post-trained from released weights on the same windows for each embodiment.
For all baselines and \method{} variants, they share the same training dataset, schedule, and initializations. 
We fire the \method{} term every $8$ iterations and set $\lambda$ so that it is $15\%$ of the flow loss on a short calibration run. 
Other details are in Appendices~\ref{sec:icm-backbone}--\ref{sec:icm-training}.

\textbf{Baselines.}
We compare against competing robot world models as baselines for \method{}. 
First, iVideoGPT~\citep{wu2024ivideogpt} is an autoregressive token-based model.
CoCo~\citep{shi2026coco}, which adds counterfactual-consistency constraints to iVideoGPT, was reproduced by us as close as possible while following their paper. As noted in their paper, the ASC2 component of their paper is not suited for joint-space actions like in our settings, so we only implement the MSC2 component.
Ctrl-World~\citep{guo2026ctrlworld} is a video-diffusion world model for policy evaluation.
We also report DreamDojo and Cosmos3 trained with the flow loss alone, which isolates the effect of \method{}. 
All baselines are trained on the same perturbation-play corpora, and have a fixed budget, where they see the same amount of samples during training.

\textbf{Evaluation.}
Following Ctrl-World~\citep{guo2026ctrlworld}, we evaluate the models in a closed loop, where policies generate actions, world models predict videos conditioned on those actions, and the generated observations are fed back into the policy.
We report PSNR, LPIPS~\citep{zhang2018lpips}, and FVD~\citep{unterthiner2018fvd} for a single prediction window (short horizon) and for autoregressive rollouts chained over multiple windows (long horizon).
To measure action fidelity, we compute the Pearson correlation between real and generated rollout success rates across policy checkpoints and tasks.
Each generated video receives a binary success label from human evaluators, with success labels only given when the task is clearly succeeded in the output video and the target object and environment remain visible throughout the rollout.
For the DROID and ALOHA-2 setup, we utilize GR00T N1.5~\citep{nvidia2025gr00tn15}, FLARE~\citep{zheng2025flare}, and DUST~\citep{won2026dust} as policy models for action generation. For UR5e on the OGBench setup, we utilize GCIQL and GCIVL policies, which are both defined in OGBench~\citep{park2025ogbench}.
For each embodiment, we utilize 12 episodes per 4 checkpoints across 2 tasks.

\textbf{Simulation results.}
Table~\ref{Tab:tab_sim_ur5e} reports results on UR5e.
\method{} improves both backbones on every reported metric, including video quality at both horizons and the correlation between real and imagined policy success rates.
For DreamDojo, Pearson correlation increases from $0.451$ to $0.686$, while FVD decreases from $85$ to $73$ at the short horizon and from $108$ to $93$ at the long horizon.
DreamDojo + \method{} achieves the best score on every metric among the evaluated methods.
For Cosmos3, correlation increases from $0.091$ to $0.334$, alongside smaller FVD reductions from $470$ to $463$ and from $562$ to $540$.
These results show that the gains extend beyond single-window visual quality to policy evaluation and autoregressive rollout.

\textbf{Real robot results.}
Table~\ref{Tab:tab_real} reports results on the DROID and ALOHA-2 configurations.
On DROID, \method{} raises Pearson correlation from $0.468$ to $0.790$ for DreamDojo and from $0.425$ to $0.583$ for Cosmos3, while reducing FVD from $807$ to $622$ ($23\%$) and from $910$ to $744$ ($18\%$), respectively.
Both backbones also improve in PSNR and LPIPS.
On ALOHA-2, Cosmos3's correlation increases from $0.298$ to $0.694$, the highest among the evaluated methods, while its FVD decreases from $333$ to $257$.
DreamDojo's correlation improves from $0.325$ to $0.490$, with PSNR increasing from $24.20$ to $24.91$ and LPIPS decreasing from $0.06$ to $0.05$, although FVD rises slightly from $183$ to $188$.
Thus, \method{} consistently improves agreement with real policy success rates, while generally improving visual metrics.

\subsection{Robotics world models for planning}
\label{sec:exp-planning}
\begin{figure*}[t]
\centering\small
\includegraphics[width=\linewidth]{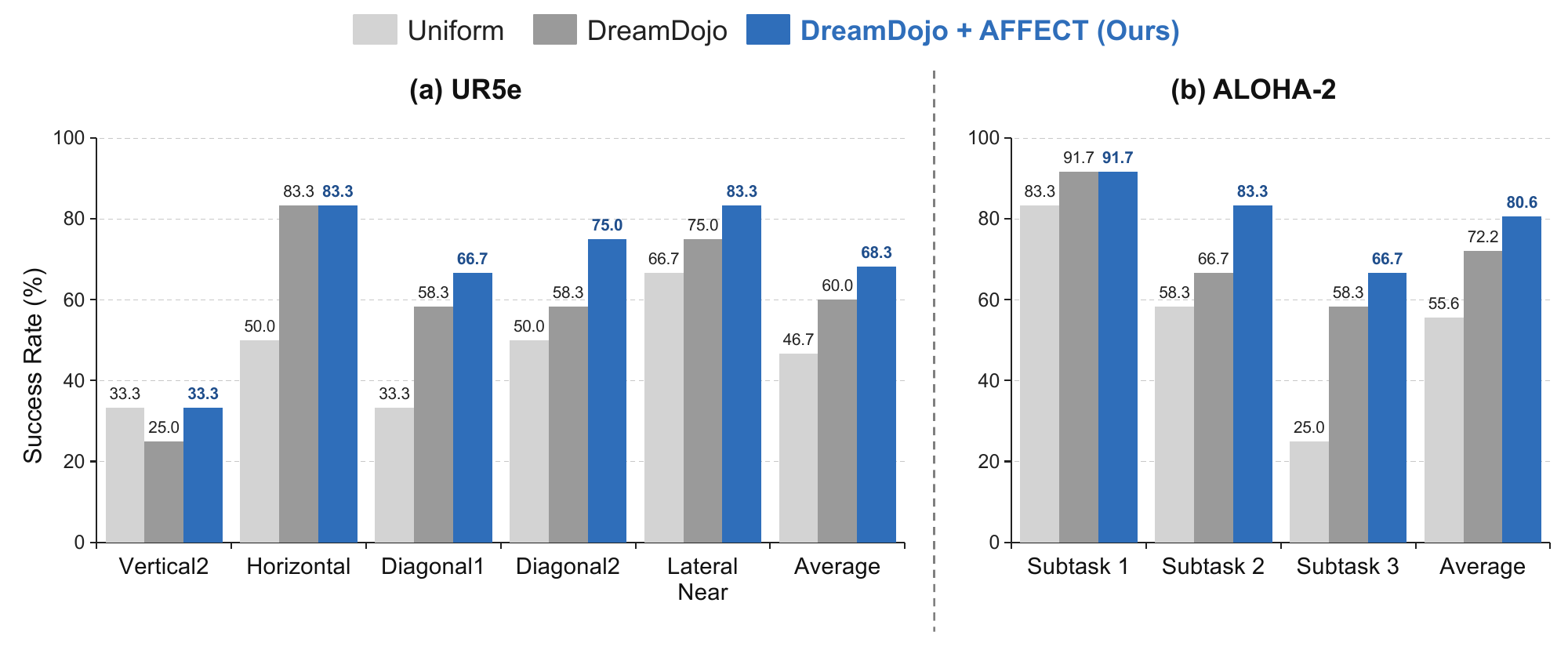}
\vspace{-15pt}
\caption{
{\bf Planning with action-faithful world models.}
Success rates (\%) of plans selected with each world model on (a) UR5e and (b) ALOHA-2, where one long-horizon task consists of three consecutive subtasks.
Candidate action chunks are proposed by an ensemble of five policy checkpoints. 
Uniform picks one at random, while DreamDojo and DreamDojo + \method{} imagine each chunk and select the one whose imagined video scores highest under a learned success estimator.
}
\label{fig:planning}
\vspace{-15pt}
\end{figure*}

\textbf{Evaluation setup.}
We test whether improved action fidelity helps select actions at test time using the proposal-selection scheme of DreamDojo~\citep{gao2026dreamdojo}.
At the start of each execution horizon, 5 policy checkpoints trained on the target task are each queried for candidate action trajectories. Using the current camera frame and the candidate actions as input, the world model imagines future video for each candidate, and a trained success estimator scores each video. 
The candidate action chunk with the highest score is executed open-loop before replanning, at which we repeat the process.
We compare DreamDojo and DreamDojo + \method{} with the same policy ensemble and success estimator. 
An additional baseline, Uniform, selects a candidate action at random rather than utilizing a world model.
The score estimator uses frozen DINOv2 features with a light transformer head to predict task-success probability given video input.
Further details are in Appendix~\ref{sec:planning-details}.

\textbf{Tasks and Results.}
For the UR5e setup, we evaluate on success rate over 5 select pick-and-place tasks from OGBench, while we use a long-horizon task for ALOHA-2, consisting of 3 consecutive tasks where the robot must open a trash bin, and throw away two items on the table.
\Figref{fig:planning} shows that average UR5e success increases from $46.7\%$ with Uniform to $60.0\%$ with DreamDojo and $68.3\%$ with DreamDojo + \method{}, an additional gain of $8.3$ percentage points.
ALOHA-2 results also show that \method{} provides a better environment for planning over the baseline world model, with an average gain of $8.4$ percentage points.
With the proposal ensemble and success estimator held fixed, the improvements show the usefulness of action-effect matching for improving model-based planning.

\subsection{Analysis and Ablations}

\begin{wrapfigure}{r}{0.40\textwidth}
\centering\small
\vspace{-12pt}
\includegraphics[width=\linewidth]{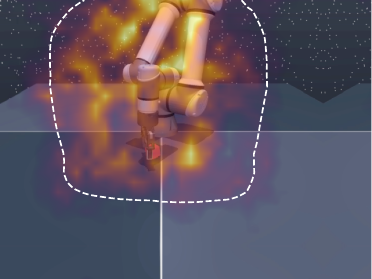}
\vspace{-16pt}
\caption{\textbf{Action-effect localization.}
Frozen-feature responses on UR5e relative to shuffled pairs, where brighter cells indicate stronger responses.}
\label{fig:effect-signal}
\vspace{-14pt}
\end{wrapfigure}

\textbf{Do the frozen features see the action effect?}
The frozen video features detect action effects primarily where the robot changes the scene (\Figref{fig:effect-signal}).
We establish this on recorded UR5e videos by measuring the consistency of perturbation--feature associations at each latent cell. We check whether products of perturbation and video features have a consistent sign across episodes, then average over feature pairs.
We also evaluate on a baseline setup that randomly pairs perturbations with videos from other windows, breaking the link between action perturbations and outcomes.
Responses are 1.6--2.2 times stronger than this shuffled baseline in regions traversed by the arm, cube, or shadow, which is where action-induced visual change is concentrated, but remain at chance on the static table and backdrop.
This confirms that the features provide a localized action-effect signal for \method{}.

\textbf{Ablation studies.}
We compare six Cosmos3 configurations after $10$k training iterations on perturbation play data, evaluated on video fidelity metrics (PSNR, LPIPS, FVD) to examine the role of action--outcome correspondence, perturbation weighting, and training-data quantity (Table~\ref{Tab:tab_ablation}).
Alongside the flow-only baseline and Cosmos3 + \method{}, we consider two controls.
\emph{Placebo matching} shuffles perturbation features $\bm{g}_t$ across episodes, isolating the importance of correct action--outcome correspondence.
\emph{Unconditional feature matching} sets $\bm{g}_t \equiv \bm{1}$ to match mean video features, isolating the contribution of perturbation weighting beyond generic feature matching.
Finally, we train the flow-only baseline and Cosmos3 + \method{} on the same $20\%$ subset of the training data to assess the method under limited data.

Results in Table~\ref{Tab:tab_ablation} show that both controls show degraded video fidelity, supporting the importance of both preserving action-outcome correspondence and using perturbation-weighted features.
Meanwhile, \method{} trained under data constraints shows a lower performance gain on its baseline compared to the full data run, showing that its performance gain increases with further action distribution coverage as expected, while still helping at lower data scales.

\begin{table}[t]
\centering
\caption{{\bf Ablations.}
We report long-horizon video metrics (PSNR, LPIPS, FVD) for Cosmos3 using the same closed-loop policy evaluation system as Table~\ref{Tab:tab_real} after $10{,}000$ training iterations, varying the auxiliary objective  and training-data quantity.}
\vspace{0.10in}
\begin{tabular}{l ccc}
\toprule
Method & PSNR $\uparrow$ & LPIPS $\downarrow$ & FVD $\downarrow$ \\
\midrule
Cosmos3 & 23.43 & 0.07 & 284 \\
w/ Placebo matching & 21.80 & 0.09 & 460 \\
w/ Unconditional matching & 21.72 & 0.09 & 451 \\
{\bf w/ \method{} (Ours)} & {\bf 24.12} & {\bf 0.06} & {\bf 256} \\
\midrule
\multicolumn{4}{l}{\emph{20\% training data}} \\
Cosmos3 & 22.58 & 0.08 & 405 \\
{\bf w/ \method{} (Ours)} & 22.97 & 0.08 & 393 \\
\bottomrule
\end{tabular}
\label{Tab:tab_ablation}
\end{table}

\section{Related works}
\textbf{Robotics world models (RWMs).}
Learned world models began as compact latent dynamics models used for planning or actor-critic learning in imagination \citep{ha2018world,hafner2019planet,hafner2025dreamerv3,hansen2024tdmpc2}, and have since become large video generators conditioned on actions or text \citep{yang2024unisim,du2023unipi,bruce2024genie,parkerholder2025genie3,valevski2025gamengen,alonso2024diamond}.
For manipulation, the dominant recipe is to post-train a general pretrained video diffusion or flow model with action conditioning \citep{zhu2024irasim,wu2024ivideogpt,huang2026vid2world,guo2026ctrlworld}.
These models are used to generate data \citep{jang2025dreamgen}, to plan \citep{du2024vlp,bar2025nwm}, and increasingly to evaluate policies without actual real-world rollout \citep{li2025worldeval,jeon2026roboworld,ma2026gigaworld}. 

\textbf{Action fidelity in world models.}
A growing body of mostly concurrent work finds that RWMs often render plausible videos that ignore their action condition.
Benchmarks pair successful and failure-inducing actions from a fixed scene \citep{yang2026mirabench,shi2026coco} or score robot-object interaction directly \citep{jiang2026robowmbench}, and position papers argue that counterfactual fidelity rather than visual quality should be the deciding criterion \citep{wang2026counterfactualcontrol,yu2026decisioncentric}.
Inverse-dynamics probes \citep{li2025worldeval,wang2026eva,yeom2026actionrelevant} test action recoverability from generated video, and similar ideas have been turned into training objectives that reconstruct the action or separate predictions made under different actions \citep{ye2025reinforcement,han2026wam,gan2026actswm}.
Because these objectives can be satisfied by distinguishable robot motion alone, they do not guarantee correct robot-object interaction.
Other methods supervise outcomes directly, through replayed off-expert actions in simulation \citep{chen2026worldecho} or through additional outcome labels and rewards \citep{zhou2026clwm,wang2026pavxplorerl}, which limits how far they scale on real robots.
\method{} needs neither paired counterfactual rollouts nor outcome labels.

\section{Conclusion}
\label{sec:conclusion}
We presented \method{}, an auxiliary training objective that trains robot world models to learn how action perturbations change future observations.
Using perturbation play rollouts with logged perturbations and action effect encoders to extract video/perturbation features, it adds action effect supervision without outcome labels, repeated executions from the same state, or architectural changes.
Across DreamDojo and Cosmos3 backbones on UR5e, DROID, and ALOHA-2 hardware, \method{} improves the correlation between real and imagined policy success rates, with improvements in most visual metrics.
The additional model-based planning results further suggest that these gains can translate into better action selection.

\subsection*{AI use statement}
In this work, we used generative AI tools for editing and proofreading the manuscript text, including fixing typographical errors and checking the LaTeX source against the formatting instructions, and upscaling image resolutions for showcasing task settings.
We have reviewed all AI-assisted work, and every suggested edit was checked against the source and results by the authors.
We take responsibility for the final content of this work, including text, claims, and artifacts produced with the aid of generative AI.

\subsection*{Ethics statement}
This work trains video world models for robot manipulation and uses them to evaluate robot policies.
All real-robot data were collected by the authors on a Franka Research 3 and a bimanual ALOHA-2 in a laboratory setting, using the perturbation-play protocol of Section~\ref{sec:method-data}. Human involvement was limited to teleoperated play episodes used to fine-tune the data-collection policy and to resetting objects that left the workspace (Appendix~\ref{sec:icm-data}).
We are not aware of other ethical concerns specific to this work.

\subsection*{Reproducibility statement}
Section~\ref{sec:method} and Appendix~\ref{sec:icm-appendix} specify \method{} in full. The data-collection protocol and per-robot configurations are in Appendix~\ref{sec:icm-data}. The backbones, their action inputs, and warm starts are in Appendix~\ref{sec:icm-backbone}. The feature bank, estimator, scales, sampler, and training schedule are in Appendices~\ref{sec:icm-bank}--\ref{sec:icm-training}. 
The planning setup is in Appendix~\ref{sec:planning-details}.
Appendix~\ref{sec:icm-rationale} states the assumptions behind the objective.
Baselines and \method{} variants share data, initialization, and schedule, differing only in the auxiliary loss (Section~\ref{sec:exp-setup}).

\subsection*{Acknowledgements}
This paper was supported by RLWRLD Inc. This research was supported by the ``Advanced GPU Utilization Support Program'' funded by the Government of the Republic of Korea (Ministry of Science and ICT). We thank Jimin Lee and Byungjoon Yoon for their help in collecting robot demonstrations for our real-world experiments.

\bibliography{main_ref}

@String(CVPR  = {IEEE Conference on Computer Vision and Pattern Recognition})

@String(ICML = {International Conference on Machine Learning})

@String(ICLR = {International Conference on Learning Representations})

@String(NeurIPS = {Advances in Neural Information Processing Systems})

@String(RSS = {Robotics: Science and Systems})

@String(CORL = {Conference on Robot Learning})

@inproceedings{zhao2023learning,
    title={Learning Fine-Grained Bimanual Manipulation with Low-Cost Hardware}, 
    author={Tony Z. Zhao and Vikash Kumar and Sergey Levine and Chelsea Finn},
    year={2023},
    booktitle=RSS,
}

@article{aldaco2024aloha2,
  title={Aloha 2: An enhanced low-cost hardware for bimanual teleoperation},
  author={Aldaco, Jorge and Armstrong, Travis and Baruch, Robert and Bingham, Jeff and Chan, Sanky and Draper, Kenneth and Dwibedi, Debidatta and Finn, Chelsea and Florence, Pete and Goodrich, Spencer and others},
  journal={arXiv preprint arXiv:2405.02292},
  year={2024}
}

@inproceedings{park2025ogbench,
    title={OGBench: Benchmarking Offline Goal-Conditioned RL},
    author={Park, Seohong and Frans, Kevin and Eysenbach, Benjamin and Levine, Sergey},
    booktitle=ICLR,
    year={2025},
}

@article{khazatsky2024droid,
  title={Droid: A large-scale in-the-wild robot manipulation dataset},
  author={Khazatsky, Alexander and Pertsch, Karl and Nair, Suraj and Balakrishna, Ashwin and Dasari, Sudeep and Karamcheti, Siddharth and Nasiriany, Soroush and Srirama, Mohan Kumar and Chen, Lawrence Yunliang and Ellis, Kirsty and others},
  journal={arXiv preprint arXiv:2403.12945},
  year={2024}
}

@misc{nvidia2025gr00tn15,
    title={{GR00T N1.5}: An Improved Open Foundation Model for Generalist Humanoid Robots},
    author={{NVIDIA GEAR}},
    howpublished = {\url{https://research.nvidia.com/labs/gear/gr00t-n1_5}},
    year = {2025},
}

@article{zheng2025flare,
  title={Flare: Robot learning with implicit world modeling},
  author={Zheng, Ruijie and Wang, Jing and Reed, Scott and Bjorck, Johan and Fang, Yu and Hu, Fengyuan and Jang, Joel and Kundalia, Kaushil and Lin, Zongyu and Magne, Loic and others},
  journal={arXiv preprint arXiv:2505.15659},
  year={2025}
}

@inproceedings{won2026dust,
  title={Dual-Stream Diffusion for World-Model Augmented Vision-Language-Action Model},
  author={Won, John and Lee, Kyungmin and Jang, Huiwon and Kim, Dongyoung and Shin, Jinwoo},
  booktitle=ICML,
  year={2026}
}

@article{agarwal2026cosmos,
  title={Cosmos 3: Omnimodal world models for physical ai},
  author={Agarwal, Niket and Ali, Arslan and Allen, Jon and Antolini, Martin and Aubame, Adeline and Azzolini, Alisson and Bai, Junjie and Bala, Maciej and Balaji, Yogesh and Bapst, Josh and others},
  journal={arXiv preprint arXiv:2606.02800},
  year={2026}
}

@inproceedings{gao2026dreamdojo,
  title={Dreamdojo: A generalist robot world model from large-scale human videos},
  author={Gao, Shenyuan and Liang, William and Zheng, Kaiyuan and Malik, Ayaan and Ye, Seonghyeon and Yu, Sihyun and Tseng, Wei-Cheng and Dong, Yuzhu and Mo, Kaichun and Lin, Chen-Hsuan and others},
  booktitle=ICML,
  year={2026}
}

@inproceedings{wu2024ivideogpt,
    title={iVideoGPT: Interactive VideoGPTs are Scalable World Models}, 
    author={Jialong Wu and Shaofeng Yin and Ningya Feng and Xu He and Dong Li and Jianye Hao and Mingsheng Long},
    booktitle=NeurIPS,
    year={2024},
}

@inproceedings{guo2026ctrlworld,
    title={Ctrl-World: A Controllable Generative World Model for Robot Manipulation}, 
    author={Yanjiang Guo and Lucy Xiaoyang Shi and Jianyu Chen and Chelsea Finn},
    booktitle=ICLR,
    year={2026}
}

@article{shi2026coco,
    title={Overcoming Statistical Bias in Action-Controllable World Models}, 
    author={Yuhong Shi and Zhenhao Chu and Jie Wei and Jun Hao and Jianyi Liu and Jingwen Fu},
    year={2026},
    journal={arXiv preprint arXiv:2608.04653},
}

@inproceedings{
    lipman2023flow,
    title={Flow Matching for Generative Modeling},
    author={Yaron Lipman and Ricky T. Q. Chen and Heli Ben-Hamu and Maximilian Nickel and Matthew Le},
    booktitle=ICLR,
    year={2023}
}

@inproceedings{ha2018world,
  title={Recurrent World Models Facilitate Policy Evolution},
  author={Ha, David and Schmidhuber, J{\"u}rgen},
  booktitle=NeurIPS,
  year={2018}
}

@inproceedings{yang2024unisim,
  title={Learning Interactive Real-World Simulators},
  author={Yang, Mengjiao and Du, Yilun and Ghasemipour, Kamyar and Tompson, Jonathan and Schuurmans, Dale and Abbeel, Pieter},
  booktitle=ICLR,
  year={2024}
}

@inproceedings{bruce2024genie,
  title={Genie: Generative interactive environments},
  author={Bruce, Jake and Dennis, Michael D and Edwards, Ashley and Parker-Holder, Jack and Shi, Yuge and Hughes, Edward and Lai, Matthew and Mavalankar, Aditi and Steigerwald, Richie and Apps, Chris and others},
  booktitle=ICML,
  year={2024}
}

@article{zhu2024irasim,
    title={IRASim: Learning Interactive Real-Robot Action Simulators},
    author={Fangqi Zhu and Hongtao Wu and Song Guo and Yuxiao Liu and Chilam Cheang and Tao Kong},
    journal={arXiv preprint arXiv:2406.14540},
    year={2024}
}

@inproceedings{hafner2019planet,
  title={Learning Latent Dynamics for Planning from Pixels},
  author={Hafner, Danijar and Lillicrap, Timothy and Fischer, Ian and Villegas, Ruben and Ha, David and Lee, Honglak and Davidson, James},
  booktitle=ICML,
  year={2019}
}

@article{hafner2025dreamerv3,
  title={Mastering diverse control tasks through world models},
  author={Hafner, Danijar and Pasukonis, Jurgis and Ba, Jimmy and Lillicrap, Timothy},
  journal={Nature},
  volume={640},
  number={8059},
  pages={647--653},
  year={2025},
  publisher={Nature Publishing Group UK London}
}

@inproceedings{hansen2024tdmpc2,
  title={{TD-MPC2}: Scalable, Robust World Models for Continuous Control},
  author={Hansen, Nick and Su, Hao and Wang, Xiaolong},
  booktitle=ICLR,
  year={2024}
}

@inproceedings{du2023unipi,
  title={Learning universal policies via text-guided video generation},
  author={Du, Yilun and Yang, Sherry and Dai, Bo and Dai, Hanjun and Nachum, Ofir and Tenenbaum, Josh and Schuurmans, Dale and Abbeel, Pieter},
  booktitle=NeurIPS,
  year={2023}
}

@inproceedings{du2024vlp,
  title={Video language planning},
  author={Du, Yilun and Yang, Sherry and Florence, Pete and Xia, Fei and Wahid, Ayzaan and Sermanet, Pierre and Yu, Tianhe and Abbeel, Pieter and Tenenbaum, Joshua B and Kaelbling, Leslie and others},
  booktitle=ICLR,
  year={2024}
}

@misc{parkerholder2025genie3,
  title={{Genie 3}: A New Frontier for World Models},
  author={Parker-Holder, Jack and Fruchter, Shlomi},
  howpublished={Google DeepMind blog, \url{https://deepmind.google/blog/genie-3-a-new-frontier-for-world-models/}},
  year={2025}
}

@inproceedings{valevski2025gamengen,
  title={Diffusion models are real-time game engines},
  author={Valevski, Dani and Leviathan, Yaniv and Arar, Moab and Fruchter, Shlomi},
  booktitle=ICLR,
  year={2025}
}

@inproceedings{alonso2024diamond,
  title={Diffusion for world modeling: Visual details matter in atari},
  author={Alonso, Eloi and Jelley, Adam and Micheli, Vincent and Kanervisto, Anssi and Storkey, Amos and Pearce, Tim and Fleuret, Fran{\c{c}}ois},
  booktitle=NeurIPS,
  year={2024}
}

@inproceedings{bar2025nwm,
  title={Navigation World Models},
  author={Bar, Amir and Zhou, Gaoyue and Tran, Danny and Darrell, Trevor and LeCun, Yann},
  booktitle=CVPR,
  year={2025}
}

@inproceedings{huang2026vid2world,
  title={Vid2world: Crafting video diffusion models to interactive world models},
  author={Huang, Siqiao and Wu, Jialong and Zhou, Qixing and Miao, Shangchen and Long, Mingsheng},
  booktitle=ICLR,
  year={2026}
}

@article{ali2025cosmospredict25,
  title={World simulation with video foundation models for physical ai},
  author={Ali, Arslan and Bai, Junjie and Bala, Maciej and Balaji, Yogesh and Blakeman, Aaron and Cai, Tiffany and Cao, Jiaxin and Cao, Tianshi and Cha, Elizabeth and Chao, Yu-Wei and others},
  journal={arXiv preprint arXiv:2511.00062},
  year={2025}
}

@inproceedings{jang2025dreamgen,
  title={Dreamgen: Unlocking generalization in robot learning through video world models},
  author={Jang, Joel and Ye, Seonghyeon and Lin, Zongyu and Xiang, Jiannan and Bjorck, Johan and Fang, Yu and Hu, Fengyuan and Huang, Spencer and Kundalia, Kaushil and Lin, Yen-Chen and others},
  booktitle=CORL,
  year={2025}
}

@article{li2025worldeval,
  title={Worldeval: World model as real-world robot policies evaluator},
  author={Li, Yaxuan and Zhu, Yichen and Wen, Junjie and Shen, Chaomin and Xu, Yi},
  journal={arXiv preprint arXiv:2505.19017},
  year={2025}
}

@article{jeon2026roboworld,
  title={RoboWorld: Fast and Reliable Neural Simulators for Generalist Robot Policy Evaluation},
  author={Jeon, Byeongguk and Ye, Seonghyeon and Doo, JaeHyeok and Kim, Sungdong and Seo, Minjoon and Son, Hyungmok and Lee, Kimin},
  journal={arXiv preprint arXiv:2607.01060},
  year={2026}
}

@article{ma2026gigaworld,
  title={GigaWorld-1: A Roadmap to Build World Models for Robot Policy Evaluation},
  author={Team, GigaWorld and Ma, Angyuan and Wang, Boyuan and Li, Bohan and Ni, Chaojun and Li, Guo and Huang, Guan and Zhao, Guosheng and Li, Hao and Li, Hengtao and others},
  journal={arXiv preprint arXiv:2607.02642},
  year={2026}
}

@article{yang2026mirabench,
  title={MiraBench: Evaluating Action-Conditioned Reliability in Robotic World Models},
  author={Yang, Tianzhuo and Shen, Zihan and Mi, Zirui and Zhang, Zhaoyi and Zhou, Jiayi and Ji, Jiaming and Dai, Juntao and Chen, Jiawei and Chen, Boyuan and Yang, Yaodong},
  journal={arXiv preprint arXiv:2605.29360},
  year={2026}
}

@article{wang2026counterfactualcontrol,
  title={Autonomous Video Generation with Counterfactual Controllability for Self-Evolving World Models},
  author={Wang, Xin and Liu, Wenxuan and Feng, Tongtong and Zhu, Wenwu},
  journal={arXiv preprint arXiv:2606.24152},
  year={2026}
}

@article{chen2026worldecho,
  title={Do Robotic World Models Really Follow Actions? Diagnosing and Aligning Action-Conditioned Generation for Policy Learning},
  author={Chen, Sixiang and Liu, Jiaming and Wu, Jixian and Guo, Yichen and Wang, Tinghao and Qian, Siyuan and Chen, Hao and Cao, Jiajun and Tang, Jian and Zhang, Shanghang},
  journal={arXiv preprint arXiv:2608.24885},
  year={2026}
}

@article{jiang2026robowmbench,
  title={Robowm-bench: A benchmark for evaluating world models in robotic manipulation},
  author={Jiang, Feng and Chen, Yang and Xu, Kyle and Liu, Yuchen and Wang, Haifeng and Shen, Zhenhao and Lu, Jasper and Huang, Shengze and Wang, Yuanfei and Xie, Chen and others},
  journal={arXiv preprint arXiv:2604.19092},
  year={2026}
}

@inproceedings{lynch2019play,
  title={Learning Latent Plans from Play},
  author={Lynch, Corey and Khansari, Mohi and Xiao, Ted and Kumar, Vikash and Tompson, Jonathan and Levine, Sergey and Sermanet, Pierre},
  booktitle=CORL,
  year={2020}
}

@article{yin2026playworld,
  title={Playworld: Learning robot world models from autonomous play},
  author={Yin, Tenny and Mei, Zhiting and Zheng, Zhonghe and Yamane, Miyu and Wang, David and Sceats, Jade and Bateman, Samuel M and Zha, Lihan and Badithela, Apurva and Shorinwa, Ola and others},
  journal={arXiv preprint arXiv:2603.09030},
  year={2026}
}

@inproceedings{zhang2018lpips,
  title={The Unreasonable Effectiveness of Deep Features as a Perceptual Metric},
  author={Zhang, Richard and Isola, Phillip and Efros, Alexei A. and Shechtman, Eli and Wang, Oliver},
  booktitle=CVPR,
  year={2018}
}

@article{unterthiner2018fvd,
  title={Towards accurate generative models of video: A new metric \& challenges},
  author={Unterthiner, Thomas and Van Steenkiste, Sjoerd and Kurach, Karol and Marinier, Raphael and Michalski, Marcin and Gelly, Sylvain},
  journal={arXiv preprint arXiv:1812.01717},
  year={2018}
}

@article{yu2026decisioncentric,
  title={How should world models be evaluated for embodied decision-making? A decision-making-centric position},
  author={Yu, Yang and Zhang, Shiyuan and Sheng, Yifei and Ren, Haoxiang and Lin, Haoxin},
  journal={arXiv preprint arXiv:2606.15032},
  year={2026}
}

@article{wang2026eva,
  title={Eva: Aligning video world models with executable robot actions via inverse dynamics rewards},
  author={Wang, Ruixiang and Liu, Qingming and Deng, Yueci and Liu, Guiliang and Liu, Zhen and Jia, Kui},
  journal={arXiv preprint arXiv:2603.17808},
  year={2026}
}

@article{yeom2026actionrelevant,
  title={What Makes Video World Model Latents Action-Relevant: Prediction over Reconstruction},
  author={Yeom, Jewon and Kim, Hanseul and Park, Jeongjae and Jung, Sungmok and Lee, Jaejin and Kim, Taesup},
  journal={arXiv preprint arXiv:2606.07687},
  year={2026}
}

@article{gretton2012mmd,
  title={A kernel two-sample test},
  author={Gretton, Arthur and Borgwardt, Karsten M and Rasch, Malte J and Sch{\"o}lkopf, Bernhard and Smola, Alexander},
  journal={The journal of machine learning research},
  volume={13},
  pages={723--773},
  year={2012},
  publisher={JMLR. org}
}

@inproceedings{li2015gmmn,
  title={Generative Moment Matching Networks},
  author={Li, Yujia and Swersky, Kevin and Zemel, Richard},
  booktitle=ICML,
  year={2015}
}

@inproceedings{dziugaite2015mmdopt,
  title={Training Generative Neural Networks via Maximum Mean Discrepancy Optimization},
  author={Dziugaite, Gintare Karolina and Roy, Daniel M. and Ghahramani, Zoubin},
  booktitle={Uncertainty in Artificial Intelligence (UAI)},
  year={2015}
}

@inproceedings{ren2016cgmmn,
  title={Conditional generative moment-matching networks},
  author={Ren, Yong and Zhu, Jun and Li, Jialian and Luo, Yucen},
  booktitle=NeurIPS,
  year={2016}
}

@article{liu2023rectified,
  title={Flow straight and fast: Learning to generate and transfer data with rectified flow},
  author={Liu, Xingchao and Gong, Chengyue and Liu, Qiang},
  journal={arXiv preprint arXiv:2209.03003},
  year={2022}
}

@article{wan2025wan,
  title={Wan: Open and advanced large-scale video generative models},
  author={Wan, Team and Wang, Ang and Ai, Baole and Wen, Bin and Mao, Chaojie and Xie, Chen-Wei and Chen, Di and Yu, Feiwu and Zhao, Haiming and Yang, Jianxiao and others},
  journal={arXiv preprint arXiv:2503.20314},
  year={2025}
}

@article{assran2025vjepa2,
  title={V-jepa 2: Self-supervised video models enable understanding, prediction and planning},
  author={Assran, Mido and Bardes, Adrien and Fan, David and Garrido, Quentin and Howes, Russell and Muckley, Matthew and Rizvi, Ammar and Roberts, Claire and Sinha, Koustuv and Zholus, Artem and others},
  journal={arXiv preprint arXiv:2506.09985},
  year={2025}
}

@inproceedings{zhao2023unipc,
  title={Unipc: A unified predictor-corrector framework for fast sampling of diffusion models},
  author={Zhao, Wenliang and Bai, Lujia and Rao, Yongming and Zhou, Jie and Lu, Jiwen},
  booktitle=NeurIPS,
  year={2023}
}

@article{ye2025reinforcement,
  title={Reinforcement learning with inverse rewards for world model post-training},
  author={Ye, Yang and He, Tianyu and Yang, Shuo and Bian, Jiang},
  journal={arXiv preprint arXiv:2509.23958},
  year={2025}
}

@article{han2026wam,
  title={Enhancing policy learning with world-action model},
  author={Han, Yuci and Yilmaz, Alper},
  journal={arXiv preprint arXiv:2603.28955},
  year={2026}
}

@article{gan2026actswm,
  title={ActSWM: Action-Sensitive World Models for Long-Horizon Planning in Open-World Games},
  author={Gan, Zhenfeng and Zeng, ZiTong and Cheng, Jiajun and Song, Yeke and Tang, Yongyi and Wang, Xueqian},
  journal={arXiv preprint arXiv:2607.26712},
  year={2026}
}

@article{zhou2026clwm,
  title={Learning Counterfactual World Models for Embodied Reasoning under Partial Observability},
  author={Zhou, Todd Y and Zhang, Daniel},
  journal={arXiv preprint arXiv:2609.05834},
  year={2026}
}

@article{wang2026pavxplorerl,
  title={PAVXploreRL: Physical-Action-Visual World Model Reinforcement Learning with Action Exploration},
  author={Wang, Han and Wang, Zijun and Xue, Shuoshuo and Cao, Rui and Cheng, Fengjiao and Liang, Xiaodan and Lee, Roy Ka-Wei},
  journal={arXiv preprint arXiv:2607.16602},
  year={2026}
}

@article{kong2024hunyuanvideo,
  title={Hunyuanvideo: A systematic framework for large video generative models},
  author={Kong, Weijie and Tian, Qi and Zhang, Zijian and Min, Rox and Dai, Zuozhuo and Zhou, Jin and Xiong, Jiangfeng and Li, Xin and Wu, Bo and Zhang, Jianwei and others},
  journal={arXiv preprint arXiv:2412.03603},
  year={2024}
}

@misc{oquab2023dinov2,
  title={DINOv2: Learning Robust Visual Features without Supervision},
  author={Oquab, Maxime and Darcet, Timothée and Moutakanni, Theo and Vo, Huy V. and Szafraniec, Marc and Khalidov, Vasil and Fernandez, Pierre and Haziza, Daniel and Massa, Francisco and El-Nouby, Alaaeldin and Howes, Russell and Huang, Po-Yao and Xu, Hu and Sharma, Vasu and Li, Shang-Wen and Galuba, Wojciech and Rabbat, Mike and Assran, Mido and Ballas, Nicolas and Synnaeve, Gabriel and Misra, Ishan and Jegou, Herve and Mairal, Julien and Labatut, Patrick and Joulin, Armand and Bojanowski, Piotr},
  journal={arXiv:2304.07193},
  year={2023}
}
\bibliographystyle{iclr2027_conference}

\appendix

\section{\method{} implementation details}
\label{sec:icm-appendix}
This appendix specifies more implementation details on \method{}. Section~\ref{sec:icm-data}
describes how the training corpus is collected. Section~\ref{sec:icm-backbone} describes the two
video backbones, and Section~\ref{sec:icm-setting} fixes the setting and the notation of the loss.
Sections~\ref{sec:icm-objective}--\ref{sec:icm-training} give the objective, the feature banks, the
estimator, the per-test scales, the sampler and the training schedule.

\subsection{Data collection by perturbation play on real robots}
\label{sec:icm-data}

\method{} is trained on \emph{perturbation play}. A policy is run closed-loop on the real robot, and
every command it issues is perturbed by noise drawn from a process we control and log. Neither
demonstrations of the target task nor paired rollouts are needed.

\paragraph{Play policy and rollouts.} The nominal behavior comes from a generalist imitation policy
fine-tuned on a small set of human \emph{play} teleoperation episodes, in which an operator interacts
freely with a handful of tabletop objects instead of repeating one task. We use GR00T-N1.5 with the
FLARE objective \citep{nvidia2025gr00tn15,zheng2025flare} on both robots. At each replanning point the
policy observes the cameras and the proprioceptive state, emits a chunk of absolute joint-position
targets, and the controller executes the first $H$ steps before the policy is queried again. Episodes
are rolled to a fixed step budget and never gated on success. Objects are left where the previous
rollout put them, and a human intervenes only to return objects that leave the workspace, so
coverage grows by drift, in the spirit of autonomous play collection \citep{yin2026playworld}.
Failed grasps, collisions and knocked-over objects are kept, since they are action-dependent
outcomes the model has to learn.

\paragraph{Noising.} Let $\bar{\bm{a}}_t \in \mathbb{R}^D$ be the nominal target at step $t$. We draw a
perturbation $\bm{u}_t \in \mathbb{R}^D$ and execute
\begin{equation}
  \bm{a}_t \;=\; \mathrm{clip}\bigl(\bar{\bm{a}}_t + \bm{u}_t\bigr),
  \label{eq:icm-exec}
\end{equation}
where $\mathrm{clip}$ enforces the joint limits and a per-step velocity limit.
Stacking $H$ consecutive steps gives the chunks $\bar{\bm{A}}_t$, $\bm{U}_t$ and $\bm{A}_t$ of Section~\ref{sec:method-data}. The noise is injected
after the policy, in the absolute joint-target coordinates the robot is commanded in, so its law is
specified independently of the policy. Within an episode the recipe is as follows.
\begin{itemize}
\item \textbf{Declared scale.} Each episode declares an amplitude $\rho$ drawn from a small ladder
  (five rungs on FR3) and a fixed per-dimension shape $\bm{c} \in \mathbb{R}^D$, so that the per-dimension
  scale is $\bm{\sigma} = \rho\, \bm{c}$. The shape is the per-dimension standard deviation of
  $\bar{\bm{a}}_t - \bar{\bm{a}}_{t-1}$ on the play corpus, normalized to mean one, and is frozen once for the
  corpus. It is not re-fitted to the realized noise, since a scale fitted to the same $\bm{u}_t$ would
  turn the magnitude tests of Section~\ref{sec:icm-bank} into constants.
\item \textbf{Temporal correlation.} Per-step white noise averages out within a chunk and only makes
  the arm jitter in place. We therefore draw white Gaussian noise for the whole episode, smooth it
  along time with a Gaussian filter of width $4$ steps, re-standardize each dimension, and multiply
  elementwise by $\bm{\sigma}$.
\item \textbf{Grippers are not perturbed.} The gripper slots of $\bm{u}_t$ are zero, because gripper noise produces failure modes that no policy produces.
\item \textbf{Independent streams.} Each episode draws from its own generator, seeded from the arm and episode indices. The $\rho = 0$ arm is the unmodified policy.
\end{itemize}

\paragraph{Logged quantities.} For every step we log the nominal command $\bar{\bm{a}}_t$, the noise $\bm{u}_t$ \emph{before clipping}, the executed command $\bm{a}_t$, the number of clipped dimensions, the joint state and the position within the chunk. 
For every episode we log $\rho$, $\bm{c}$, the smoothing width, the seed and one video per camera aligned with $\bm{a}_t$. 
The noise is logged before clipping so that it follows the specified noise process. 
Clipping is rare at our scales (on FR3, at most $10^{-4}$ clipped dimensions per step on average). 
The scale is declared rather than estimated, because one batch mixes several amplitude rungs and a batch estimate would make the centering data-dependent.
Because the declared parameters are part of $\mathcal{H}_t$, pooling rungs keeps each perturbation feature centered within every rung.

\paragraph{The two instantiations.}

\begin{center}
\small
\begin{tabular}{lp{0.36\linewidth}p{0.36\linewidth}}
\hline
 & FR3 (DROID rig) & ALOHA-2 \\
\hline
Robot & Franka Research 3, 7 joints $+$ parallel gripper \citep{khazatsky2024droid} &
  two 6-DoF arms $+$ two grippers, $D = 14$ \citep{zhao2023learning,aldaco2024aloha2} \\
Perturbed dims & $7$ arm joints (gripper zero) & $12$ arm joints (both grippers zero) \\
Cameras & two exterior $+$ wrist, $256\times192$ & high, low, two wrist, $480\times640$ \\
Control rate & $10$~Hz & $25$~Hz \\
\hline
\end{tabular}
\end{center}

\subsection{Action-conditioned rectified-flow backbones}
\label{sec:icm-backbone}

We train \method{} on two backbones that differ in almost every architectural respect. 
Both are conditional video generators post-trained from released weights, and both consume the same windows through the same loader.

\paragraph{Window and latent.} A training example is a window of $H + 1$ consecutive frames $\bm{o}_{t:t+H}$ from one camera and the $H$ executed commands $\bm{A}_t=(\bm{a}_t,\dots,\bm{a}_{t+H-1})$ between them ($H = 12$ at $10$~Hz on FR3, so a window is $1.2$~s). 
A frozen causal video tokenizer $\mathrm{Enc}$ with temporal compression $4$ maps the window to a latent $\bm{Z} = \mathrm{Enc}(\bm{o}_{t:t+H}) \in \mathbb{R}^{C \times (1 + H/4) \times h \times w}$ with latent frames $\bm{z}_0,\dots,\bm{z}_{H/4}$ indexed by $m$. 
For $H = 12$ this gives four latent frames, the first of which encodes the current frame. 
All video features in Section~\ref{sec:icm-bank} are functions of this latent, and $\bm{O}_t$ and $\hat{\bm{O}}_t$ there denote the recorded and generated latents $\bm{Z}$, including the conditioning frame $\bm{z}_0$.

\paragraph{Action input.} The loader applies a fixed operator $\Delta$ to the executed commands and passes $\Delta(\bm{A}_t) \in \mathbb{R}^{H \times D}$ to the model. 
The operator expresses each four-step chunk relative to the step before the chunk and standardizes each dimension by a corpus-level statistic. 
The two backbones then ingest the chunk differently.

\begin{center}
\small
\begin{tabular}{lp{0.36\linewidth}p{0.36\linewidth}}
\hline
 & DreamDojo (Cosmos-Predict2, 2B) & Cosmos3-Edge (4B) \\
\hline
Architecture & video DiT, flow matching \citep{gao2026dreamdojo,ali2025cosmospredict25} &
  omnimodal mixture-of-transformers with a frozen reasoning tower and a trained generation tower
  \citep{agarwal2026cosmos} \\
Latent for $480\times640$ & $16 \times 4 \times 60 \times 80$ & $48 \times 4 \times 30 \times 40$
  (Wan2.2 VAE, \citealp{wan2025wan}) \\
Frame conditioning & frame replacement, in which the first one or two latent frames are overwritten
  with the ground truth at every denoising step and excluded from the loss & causal sequence with the
  first latent frame clean and later frames generated autoregressively over the latent frames \\
Action conditioning & the whole chunk $\Delta(\bm{A}_t)$ is written into a fixed slot of a
  $384$-wide vector and passed through two MLPs whose outputs are added to the timestep
  embedding & the actions driving the transition $\bm{z}_m \to \bm{z}_{m+1}$ are projected by an
  embodiment-indexed linear map and interleaved as clean tokens before the frame they drive \\
Text & empty caption & one fixed sentence per embodiment \\
Warm start & DreamDojo GR-1 post-train & released Cosmos3-Edge \\
Trainable & all & generation tower and action I/O (reasoner frozen) \\
\hline
\end{tabular}
\end{center}

\paragraph{Objective and sampler.} Both are rectified-flow models \citep{liu2023rectified,
lipman2023flow}. With $\bm{\varepsilon} \sim \mathcal{N}(\bm{0}, \bm{I})$ and a noise level $\tau \in [0, 1]$, the
network sees $\bm{O}_t^\tau = (1 - \tau) \bm{O}_t + \tau \bm{\varepsilon}$ and regresses the velocity
\begin{equation}
  \mathcal{L}_{\mathrm{flow}} \;=\;
  \mathbb{E}_{(\bm{o}_t,\bm{A}_t,\bm{O}_t)\sim\mathcal{D},\,\tau,\,\bm{\varepsilon}}\,\bigl\| v_\theta(\bm{O}_t^\tau, \tau;\, \bm{o}_t,\bm{A}_t) - (\bm{\varepsilon} - \bm{O}_t) \bigr\|_2^2,
  \label{eq:icm-flow}
\end{equation}
where $\bm{o}_t$ contains the conditioning observations and text, and $\bm{A}_t$ is the executed command sequence, passed to the model as $\Delta(\bm{A}_t)$. 
Training draws the noise level from a time-shifted schedule with shift $\kappa = 5$, and generation integrates the same probability-flow ODE with an Euler scheme on that schedule. 
The two implementations of \method{} share all settings except the pooling grid of the full-random features (Section~\ref{sec:icm-bank}), and each baseline differs from its \method{} counterpart only in the loss.

\paragraph{Training details.}
For the play policy that was used in the FR3 and ALOHA-2 settings, we trained VLA models (GR00T-N1.5, FLARE, $\pi_0$) using 56k frames of play teleoperation data, which is around 100 mins of data. The VLA models were trained on 2 A100 GPUs with a batch-size of 32 over 40k iterations.

For the world model training, we utilized around 10 hours of perturbed play data for UR5e, while we utilized 5 hours of data for the FR3 and ALOHA-2 settings. Each episode within perturbation play is varied, as human intervention only occurred when the setting was disturbed enough that the play policy was not able to return to a stable setting. This meant an average rollout length of 10 minutes for each episode in the perturbed play dataset.

\subsection{Setting and notation}
\label{sec:icm-setting}

\method{} adds one auxiliary training term and changes nothing else.
The architecture, the model's inputs, the deployed sampler and the conditioning data all stay the same, so a trained model is evaluated exactly like its baseline.
The table below collects the quantities that the loss uses, grouped by where they come from.

\begin{center}
\begin{tabular}{llp{0.56\linewidth}}
\hline
Symbol & Name & Description \\
\hline
\multicolumn{3}{l}{\emph{Data and perturbation}} \\
$t$, $H$, $D$ & indices and sizes & Window start timestep, window length in steps, and action dimension. \\
$k$, $\ell$ & indices & Step within a window and action dimension. \\
$\bm{o}_t$ & context & The conditioning frames available to the model at the start of the window. \\
$\bm{a}_t\in\mathbb{R}^D$ & command & The executed command at step $t$. \\
$\bm{A}_t\in\mathbb{R}^{H\times D}$ & action & The executed command sequence $(\bm{a}_t,\dots,\bm{a}_{t+H-1})$, passed to the model as $\Delta(\bm{A}_t)$. \\
$\bar{\bm{A}}_t\in\mathbb{R}^{H\times D}$ & nominal action & The play policy's commands for this window, before noise. \\
$\bm{U}_t\in\mathbb{R}^{H\times D}$ & perturbation & The noise $(\bm{u}_t,\dots,\bm{u}_{t+H-1})$ assigned to the commands in this window,
  logged before clipping. \\
$\rho$, $\bm{c}$ & noise law & Episode amplitude and fixed per-dimension shape. \\
$\bm{\sigma}=\rho\,\bm{c}\in\mathbb{R}^D$ & noise scale & The declared per-dimension noise scale of the episode. \\
$\bm{O}_t$ & recorded future & The recorded video for this window, encoded to the model's latent. \\
$\mathcal{D}$ & dataset & The trajectory dataset. \\
$\mathcal{H}_t$ & pre-window info & Everything fixed before the window, namely $\bm{o}_t$ and the declared parameters of the noise law. \\
\hline
\multicolumn{3}{l}{\emph{World model}} \\
$W_\theta$, $v_\theta$ & world model & The conditional distribution $W_\theta(\cdot\mid\bm{o}_t,\bm{A}_t)$ and its velocity network. \\
$\tau$, $\bm{\varepsilon}$ & flow variables & Flow timestep in $[0,1]$ and Gaussian noise $\bm{\varepsilon}\sim\mathcal{N}(\bm{0},\bm{I})$. \\
$\hat{\bm{O}}_t$ & generated future & A future generated from fresh noise under the same $\bm{o}_t$ and
  $\bm{A}_t$. \\
\hline
\multicolumn{3}{l}{\emph{Loss}} \\
$\psi$, $\phi$ & encoders & Frozen perturbation and video feature encoders, with $K_\psi$ and $K_\phi$ outputs. \\
$\bm{g}_t\in\mathbb{R}^{K_\psi}$ & centered features & Centered perturbation features $\psi(\bm{U}_t)-\mathbb{E}[\psi(\bm{U}_t)\mid\mathcal{H}_t]$. \\
$b\in\mathcal{B}$ & test & One pair of a perturbation feature $g_b$ and a video feature $f_b$, the pair $(i,j)$ of the main text. \\
$s_b$ & scale & The fixed positive scale of test $b$. \\
$d_b$ & discrepancy & The per-window contribution of test $b$ to the loss (Eq.~\ref{eq:icm-total}). \\
$(n)$, $N_B$ & data units & Index and number of the independent data units (episodes) in a batch. \\
$\lambda$ & weight & Weight of $\mathcal{L}_{\mathrm{AFFECT}}$ in the total loss. \\
\hline
\end{tabular}
\end{center}

Two aspects of this setting deserve a comment.
The first is that the perturbation $\bm{U}_t$ is never a model input.
The model receives the executed command, as it would at deployment, and $\bm{U}_t$ is used only to build the loss.
Each example therefore carries, in addition to its frames and commands, the logged perturbation and the declared per-dimension scale $\bm{\sigma}$ of its episode.
No paired counterfactual futures are required, because the objective aggregates evidence across windows instead of differencing two futures from one state.

The second aspect is the relation between the auxiliary loss and the flow loss.
The flow loss is minimized by a model that reproduces the data distribution $p(\bm{O}_t\mid\bm{o}_t,\bm{A}_t)$, and ideally this same model would also attain zero \method{} loss.
It does so whenever $\bm{o}_t$ and $\bm{A}_t$ carry all the information about the outcome that $\bm{U}_t$ and $\mathcal{H}_t$ carry, that is, whenever
\begin{equation}
  \bm{O}_t \;\perp\!\!\!\perp\; (\bm{U}_t,\mathcal{H}_t) \;\mid\; (\bm{o}_t,\bm{A}_t).
  \label{eq:icm-sufficiency}
\end{equation}
To see why, write $m_f(\bm{o}_t,\bm{A}_t)=\mathbb{E}[f(\bm{O}_t)\mid\bm{o}_t,\bm{A}_t]$ for the mean of a video feature $f$ given the model's inputs, and let $g$ be any perturbation feature.
Under \eqref{eq:icm-sufficiency}, conditioning on $(\bm{U}_t,\mathcal{H}_t,\bm{A}_t)$ gives $\mathbb{E}[g\,f(\bm{O}_t)]=\mathbb{E}[g\,m_f(\bm{o}_t,\bm{A}_t)]$.
A future generated from the data distribution depends on the window only through $\bm{o}_t$ and $\bm{A}_t$, so $\mathbb{E}[g\,f(\hat{\bm{O}}_t)]$ equals the same quantity, and the two sides of every test agree.
In our data the condition holds only approximately.
The play policy observes all cameras and the proprioceptive state, while the world model is conditioned on its own frames.
Given $\bm{o}_t$ and $\bm{A}_t$, the perturbation still reveals the nominal command, and the nominal command reflects parts of the scene that the world model's context does not show.
The auxiliary loss can therefore shift the optimum away from that of the flow loss.

\subsection{The objective}
\label{sec:icm-objective}

The loss of Section~\ref{sec:method-matching} is a sum over pairs of one perturbation feature and one video feature.
We call each such pair a \emph{test}.
A test consists of a centered scalar perturbation feature $g$ and a fixed scalar video feature $f$, and it checks whether the action effect $\mathbb{E}[g\,f(\cdot)]$ of Section~\ref{sec:method-matching} is the same for recorded and generated futures.
The main text indexes the tests by a pair $(i,j)$.
In the appendix a single index $b\in\mathcal{B}$ with $|\mathcal{B}|=K_\psi K_\phi$ is more convenient, and the two notations correspond as follows.

\begin{center}
\begin{tabular}{lll}
\hline
Quantity & Main text & Appendix \\
\hline
Centered perturbation feature & $g_{t,i}$ & $g_b$, or $g_b(\bm{U}_t;\mathcal{H}_t)$ when the arguments matter \\
Video feature & $\phi_j$ & $f_b$ \\
Scale & $s_{ij}$ & $s_b$ \\
\hline
\end{tabular}
\end{center}

For a single window, the \emph{discrepancy} $d_b$ of test $b$ is the perturbation feature multiplied by the gap between the recorded and the generated video feature, divided by the scale.
The objective squares the expectation of each discrepancy and sums over tests,
\begin{equation}
  d_b \;=\; \frac{g_b(\bm{U}_t)}{s_b}\,\bigl[f_b(\bm{O}_t) - f_b(\hat{\bm{O}}_t)\bigr],
  \qquad
  \mathcal{L}_{\mathrm{AFFECT}} \;=\; \sum_{b \in \mathcal{B}} \bigl(\mathbb{E}[d_b]\bigr)^2,
  \qquad
  \mathcal{L} \;=\; \mathcal{L}_{\mathrm{flow}} + \lambda\,\mathcal{L}_{\mathrm{AFFECT}}.
  \label{eq:icm-total}
\end{equation}
The expectation runs over training windows and, for the generated future, over the model's sampling noise.
This is Eq.~\ref{eq:method-objective} of the main text in the single-index notation.

Two properties motivate this form.
First, the loss measures what the action changed rather than what the scene looked like.
Because $g_b$ has zero conditional mean given the pre-window information, any additive part of $f_b$ that is fixed before the window contributes nothing to $\mathbb{E}[d_b]$ (Proposition~1 in Appendix~\ref{sec:icm-analysis-centering}).
Second, the optimum is the observed action effect, not an extreme.
The expectation $\mathbb{E}[d_b]$ is the recorded action effect minus the generated one.
For a nonzero recorded effect, an attenuated and an exaggerated generated effect therefore produce discrepancies of opposite sign, and the square increases the loss on both sides.
Objectives that reward action recoverability or the separation of futures under different actions lack this property, since a larger response always helps them.
Appendix~\ref{sec:icm-rationale-response} works through both properties in a linear example.

Finally, stacking the tests into one feature map with entries $\Phi_b = g_b f_b/s_b$ shows that the loss is the squared finite-feature MMD of Section~\ref{sec:method-statistics} (Proposition~2 in Appendix~\ref{sec:icm-analysis-products}).

\subsection{The feature banks}
\label{sec:icm-bank}

The tests are all pairs of one output of the perturbation encoder $\psi$ and one output of the video encoder $\phi$ of Section~\ref{sec:method-matching}, so the bank is the full product $\{g\} \times \{f\}$.
Each encoder applies an elementwise nonlinearity $\eta$ followed by a fixed matrix $\bm{M}$, chosen to compute the statistics below.
Nothing in either encoder is learned, and the whole bank is reproduced from two integer seeds.

\paragraph{Perturbation features $g$.} Let $u_{t+k,\ell}$ be the noise at step $k\in\{0,\dots,H-1\}$ of the window in dimension $\ell$, and $\sigma_\ell$ the declared scale of that dimension.
For each perturbed dimension we use a \emph{signed} and a \emph{magnitude} feature,
\begin{equation}
  g^{\mathrm{sgn}}_\ell \;=\; \frac{1}{\sigma_\ell\sqrt{H}} \sum_{k=0}^{H-1} u_{t+k,\ell},
  \qquad
  g^{\mathrm{mag}}_\ell \;=\; \frac{1}{\sigma_\ell^2\sqrt{2H}} \Bigl( \sum_{k=0}^{H-1} u_{t+k,\ell}^2 - H \sigma_\ell^2 \Bigr).
  \label{eq:icm-g}
\end{equation}
The signed feature is the summed noise over the window.
It is positive when the commands in dimension $\ell$ were pushed up on balance and negative when they were pushed down.
The magnitude feature is the summed squared noise minus its expected value $H\sigma_\ell^2$.
It is positive when the window was perturbed more strongly than is typical for its episode and negative when it was perturbed less, whatever the direction.
Both features are already centered in the sense of Eq.~\ref{eq:method-centering}.
The signed sum has zero mean because the noise does, and $H\sigma_\ell^2$ is exactly the conditional expectation that the centering subtracts, which is available in closed form from the declared scale.

In addition to the per-dimension features, we use their counterparts pooled over all perturbed dimensions,
\begin{equation}
  g^{\mathrm{sgn}} \;=\; \frac{1}{\sqrt{HD_u}}\sum_{k,\ell} \frac{u_{t+k,\ell}}{\sigma_\ell},
  \qquad
  g^{\mathrm{mag}} \;=\; \frac{1}{\sqrt{2HD_u}}\Bigl(\sum_{k,\ell} \frac{u_{t+k,\ell}^2}{\sigma_\ell^2} - HD_u\Bigr),
  \label{eq:icm-g-pooled}
\end{equation}
where the sums run over the $D_u\le D$ perturbed dimensions ($D_u=7$ on FR3 and $12$ on ALOHA-2).
The signed features measure the directional response, and the magnitude features measure effects that depend on how large the perturbation was, which a signed feature cannot see (Appendix~\ref{sec:icm-rationale-response} gives an example).
Dividing by $\sigma_\ell$ makes the features comparable across dimensions and amplitudes.
The constants $\sqrt{H}$, $\sqrt{2H}$ and their pooled versions are the standard deviations that the respective sums would have under white Gaussian noise of unit scale.
They only set a convenient order of magnitude, since every test is rescaled again by $s_b$ (Appendix~\ref{sec:icm-scales}).
Because $\sigma_\ell=0$ in the unperturbed episodes with $\rho=0$, the features are undefined there, and the windows of these episodes do not enter the normalized tests.

\paragraph{Video features $f$.} The video features are fixed reductions of the model's own latent, of shape (channels $\times$ time $\times$ height $\times$ width), and are applied identically to the recorded and generated futures.
All families except the energy summarize how the latent changes over the window.
The last column gives the number of features per family on FR3.

\begin{center}
\begin{tabular}{lp{0.60\linewidth}r}
\hline
Family & Definition & Count \\
\hline
channel & For the first four channels, the change in the spatial mean from the first latent frame to
  the last, and to the second. & $8$ \\
projection & The same two differences, and the squared end difference, along $8$ fixed unit-norm
  random directions in the channel axis. & $24$ \\
flow & Mean, max, path length, late-window mean and spatial spread of the absolute temporal
  difference of the latent. & $5$ \\
energy & Mean squared latent value over the whole window. & $1$ \\
full-random & The latent is average-pooled to a fixed grid ($12\times16$ on DreamDojo, $15\times20$ on
  Cosmos3), the first-to-last difference is flattened and projected onto $8$ fixed unit-norm random
  directions. Both the signed projection and its square are used. & $16$ \\
\hline
\end{tabular}
\end{center}

The signed differences and projections are linear in the latent, while the squared projections, the flow statistics and the energy are not.
This distinction matters below, both for the conditioning frames and for the choice of sampler (Appendix~\ref{sec:icm-sampler}).

\paragraph{Bank size.} On FR3 there are $16$ perturbation features ($7$ signed and $7$ magnitude over the arm joints, plus the two pooled) and $54$ video features ($8$ channel, $24$ projection, $5$ flow, $1$ energy and $16$ full-random), for $16\times54=864$ tests.
The UR5e configuration has $192$ tests.

\paragraph{Conditioning frames.} Under DreamDojo's frame-replacement conditioning, the generated conditioning frames equal the recorded latent exactly, so their contributions cancel in the difference $f_b(\bm{O}_t)-f_b(\hat{\bm{O}}_t)$ for the linear video features.

\subsection{The estimator}
\label{sec:icm-estimator}

The objective squares an expectation, and during training this expectation has to be estimated from a minibatch.
Throughout this subsection we fix one test and drop the index $b$.
The obvious estimate squares the minibatch mean of $d$, but this estimate is biased.
For $N_B$ independent discrepancies $d^{(1)},\dots,d^{(N_B)}$,
\begin{equation}
  \mathbb{E}\Biggl[\Bigl(\frac{1}{N_B}\sum_{n=1}^{N_B} d^{(n)}\Bigr)^{2}\Biggr]
  \;=\; \bigl(\mathbb{E}[d]\bigr)^2 + \frac{\operatorname{Var}(d)}{N_B}.
  \label{eq:icm-plugin-bias}
\end{equation}
The extra term is harmful rather than merely inexact.
The discrepancy contains the generated feature $f(\hat{\bm{O}}_t)$, so part of $\operatorname{Var}(d)$ is the variability of the generated futures, and the model can lower this part by generating less diverse futures.
Minimizing the squared batch mean therefore pushes toward mode collapse, and the push is strongest for small $N_B$.

We instead average the products of distinct units,
\begin{equation}
  \widehat{\bigl(\mathbb{E}[d]\bigr)^2}
  \;=\; \frac{1}{N_B(N_B-1)}\sum_{n\neq m} d^{(n)}d^{(m)}
  \;=\; \frac{\bigl(\sum_{n=1}^{N_B} d^{(n)}\bigr)^2 - \sum_{n=1}^{N_B} \bigl(d^{(n)}\bigr)^2}{N_B(N_B-1)} .
  \label{eq:icm-ustat}
\end{equation}
Each product of two different units has expectation $\mathbb{E}[d^{(n)}]\,\mathbb{E}[d^{(m)}]=(\mathbb{E}[d])^2$ by independence, so this U-statistic is unbiased.
The second form shows that it only needs the sum and the sum of squares of the batch.
Leaving out the diagonal products $(d^{(n)})^2$ is what removes the variance penalty, although it does not by itself guarantee sample diversity.
A single minibatch estimate can also be negative, even though the quantity it estimates is not.
Summed over tests, \eqref{eq:icm-ustat} is the unbiased squared-MMD estimator of \citet{gretton2012mmd} for the feature map $\Phi$, in its paired form where each recorded future is paired with the future generated for the same window.

Unbiasedness relies on the independence of the $d^{(n)}$, so the index $n$ has to run over independent data units rather than over the rows of the batch.
Windows from one episode are dependent, and in our paired corpora the amplitude rungs of one family share a noise seed.
If two dependent rows were treated as separate units, their product would have expectation $(\mathbb{E}[d])^2$ plus their covariance, and the estimator would be biased again.
The estimator therefore sums $d$ within an independent unit of data, which is a single rollout episode in our case, before applying \eqref{eq:icm-ustat}.

\subsection{Per-test scales}
\label{sec:icm-scales}

Each test is divided by a constant $s_b$, the standard deviation of the product $g_b\,f_b(\bm{O}_t)$ over recorded windows.
Without this division the tests would enter the sum with very different magnitudes, and those with large raw values would dominate the loss.
The constant $s_b$ is the scale of the test under the recorded data alone, so it can be computed before any model is trained.
We compute it with a pass of the ordinary training loop at $\lambda=0$, so the scales see exactly the windows, window length and weighting used in training, and we freeze them before \method{} training.

The scales are frozen for a reason.
A scale recomputed in every batch would depend on the same windows as the discrepancies it divides, which would change the estimator of Appendix~\ref{sec:icm-estimator} and its expectation.
Fixed scales, in contrast, change the relative weight of the tests but not the set of models with zero loss, since $\mathbb{E}[d_b]=0$ does not depend on the value of a positive constant $s_b$.
The video features are fixed for a similar reason.
If they were optimized jointly with the generator, the choice $f_b\equiv0$ would make the loss zero for any generator.
Finally, tests with zero spread on the recorded data are dropped rather than floored.

\subsection{A differentiable sampler for \texorpdfstring{$\hat{\bm{O}}_t$}{the generated future}}
\label{sec:icm-sampler}

The loss needs a generated future $\hat{\bm{O}}_t$ with two properties.
It must be a sample generated from fresh noise, and the gradient must reach the network through generation.
We obtain both with an Euler discretization of the probability-flow ODE the model is trained on.
With time shift $\kappa$ and $S$ steps, the sampler visits the noise levels
\begin{equation}
  \tau'_k = \frac{\kappa\,\tau_k}{1 + (\kappa - 1)\tau_k},
  \qquad \tau_k = 1 - \tfrac{k}{S},\quad k = 0, \dots, S,
  \label{eq:icm-schedule}
\end{equation}
which decrease from $\tau'_0=1$ (pure noise) to $\tau'_S=0$ (clean latent).
Starting from $\hat{\bm{O}}_t^{\tau'_0}\sim\mathcal{N}(\bm{0},\bm{I})$, each step follows the predicted velocity to the next noise level,
\begin{equation}
  \hat{\bm{O}}_t^{\tau'_{k+1}} \;=\; \hat{\bm{O}}_t^{\tau'_k} + (\tau'_{k+1} - \tau'_k)\, v_\theta\bigl(\hat{\bm{O}}_t^{\tau'_k}, \tau'_k;\, \bm{o}_t,\bm{A}_t\bigr),
  \qquad k = 0, \dots, S-1,
  \label{eq:icm-euler}
\end{equation}
and the last iterate is the generated future, $\hat{\bm{O}}_t = \hat{\bm{O}}_t^{\tau'_S}$.
We use $S = 2$ and $\kappa = 5$, matching the training schedule and the deployed sampler, so the sampler visits the noise levels $1$, $5/6$ and $0$.
We checkpoint the gradient of each network call.
The recorded future is detached, so the gradient of the term flows only through the generated side.

Two cheaper ways of obtaining a generated future are tempting, but neither is a valid substitute.
The first reuses the forward pass of the flow loss and takes the denoiser's clean prediction from the training-noised latent $\bm{O}_t^\tau$.
This latent is built from the recorded $\bm{O}_t$, so the discrepancy vanishes as $\tau \to 0$ whether or not the action is used.
The second takes a single Euler step from pure noise.
At $\tau=1$ the input carries no information about $\bm{O}_t$, so the velocity that minimizes \eqref{eq:icm-flow} there is $\bm{\varepsilon}-\mathbb{E}[\bm{O}_t\mid \bm{o}_t,\bm{A}_t]$, and a full step from $\bm{\varepsilon}$ returns the conditional mean $\mathbb{E}[\bm{O}_t\mid \bm{o}_t,\bm{A}_t]$ at the population optimum.
This output is the same for every noise draw, so it is an average future rather than a sample, and it has no sampling variance.
The linear video features cannot tell the difference, but the squared ones can.
For a scalar projection $y$ of the latent, $\mathbb{E}[y^2\mid \bm{o}_t,\bm{A}_t]=(\mathbb{E}[y\mid \bm{o}_t,\bm{A}_t])^2+\operatorname{Var}(y\mid \bm{o}_t,\bm{A}_t)$, and the one-step prediction reproduces only the first term.
It would therefore match mean effects rather than distributions.

\subsection{Integration into training}
\label{sec:icm-training}

Putting the pieces together, one application of the term proceeds as follows.
\begin{enumerate}[leftmargin=*,itemsep=1pt,topsep=3pt]
\item Take the $4$ examples of the micro-batch, each with its logged perturbation $\bm{U}_t$ and declared scale $\bm{\sigma}$.
\item Generate $\hat{\bm{O}}_t$ for each example with the sampler of Appendix~\ref{sec:icm-sampler}, from fresh noise and conditioned on $\bm{o}_t$ and $\bm{A}_t$ only.
\item Compute the perturbation features from $\bm{U}_t$ and the video features of both the recorded and the generated future (Appendix~\ref{sec:icm-bank}).
\item Form the discrepancy $d_b$ of every test with the frozen scales (Appendix~\ref{sec:icm-scales}), and sum the discrepancies within each episode.
\item Apply the unbiased estimator of \eqref{eq:icm-ustat} to each test and sum over tests (Appendix~\ref{sec:icm-estimator}).
\item Add $\lambda$ times this sum to the flow loss, which is unchanged.
\end{enumerate}
The term is applied every $8$ iterations, because each application costs $S$ extra forward passes.

\subsection{Planning protocol and success estimator}
\label{sec:planning-details}

\paragraph{Proposal generation.}
Following the proposal-selection scheme of DreamDojo~\citep{gao2026dreamdojo}, we use five checkpoints from one policy training run to provide diverse action proposals.
On UR5e, these are goal-conditioned, state-based GCIQL policies trained on a play dataset of OGBench~\citep{park2025ogbench}.
For each candidate, the world model predicts $24$ future frames; the highest-scoring chunk is executed open-loop, and selection repeats from the next observation.

\paragraph{Success estimator.}
DreamDojo uses a head on frozen DINOv2 features to regress the number of steps to the next subtask boundary.
We retain the architecture but predict task-success probability, removing the need for subtask annotations.
Frames are encoded by frozen DINOv2-base~\citep{oquab2023dinov2}.
Each frame contributes $13$ tokens: the CLS token and the latent patch grid average-pooled to $3\times4$.
A clip contains four frames sampled every second frame from the $10$~Hz feature stream.
Features are projected to $256$ dimensions with token and time embeddings, and a learned task embedding is added to a summary token.
A two-layer transformer maps each clip to a success logit.
The chunk score is the mean clip success probability over the $25$-frame sequence comprising the current frame and $24$ predicted frames.

\paragraph{Comparisons.}
DreamDojo and DreamDojo + \method{} share the proposal ensemble and success estimator, so the comparison changes only the world model used to imagine candidate futures.
Uniform selects one proposal uniformly at random and serves as the ensemble reference without world-model selection.
The results are consistent with imagined futures providing a more useful distinction among candidate actions when trained with action-effect matching.

\section{Rationale for action-effect matching}
\label{sec:icm-rationale}
This appendix explains why \method{} multiplies centered perturbation features by video features, and what the resulting loss does and does not constrain.
It moves from intuition to formal statements.
Appendix~\ref{sec:icm-rationale-moments} shows that each product measures a covariance between the perturbation and the future.
Appendix~\ref{sec:icm-rationale-response} works through a simple response model to show what matching these covariances enforces.
Appendix~\ref{sec:icm-analysis} then states the formal results with their assumptions, together with an example showing why products are needed.
One caveat applies throughout.
\method{} only constrains responses that the chosen features can detect, not every physically relevant response.

\subsection{Each product measures a covariance}
\label{sec:icm-rationale-moments}

We use the notation of Section~\ref{sec:method} and Appendix~\ref{sec:icm-setting}.
Recall that the conditioning set $\mathcal{H}_t$ collects everything fixed before the window, namely the context frames $\bm{o}_t$ and the parameters of the perturbation law, such as the declared noise scale.
It does not contain the perturbation $\bm{U}_t$, the executed action $\bm{A}_t$, or the future.
Consider a single test, made of a scalar perturbation feature $g=g(\bm{U}_t;\mathcal{H}_t)$, centered so that $\mathbb{E}[g\mid\mathcal{H}_t]=0$, and a fixed scalar video feature $f$.
By the definition of the conditional covariance,
\begin{equation}
  \operatorname{Cov}(g,f(\bm{O}_t)\mid\mathcal{H}_t)
  = \mathbb{E}[g\,f(\bm{O}_t)\mid\mathcal{H}_t]
    - \underbrace{\mathbb{E}[g\mid\mathcal{H}_t]}_{=\,0}\,\mathbb{E}[f(\bm{O}_t)\mid\mathcal{H}_t]
  = \mathbb{E}[g\,f(\bm{O}_t)\mid\mathcal{H}_t],
  \label{eq:icm-rationale-covariance}
\end{equation}
so the expected product of a centered feature with a video feature is their covariance within a context.
The same identity holds for the generated future.
Subtracting the two and averaging over contexts gives
\begin{equation}
  \mathbb{E}\!\left[g\{f(\bm{O}_t)-f(\hat{\bm{O}}_t)\}\right]
  = \mathbb{E}\!\left[
      \operatorname{Cov}(g,f(\bm{O}_t)\mid\mathcal{H}_t)
      -\operatorname{Cov}(g,f(\hat{\bm{O}}_t)\mid\mathcal{H}_t)
    \right].
  \label{eq:icm-rationale-covariance-error}
\end{equation}
The left side is the quantity that the loss squares, up to the scale $s_b$.
In words, each test asks whether the video feature moves with the perturbation equally strongly in recorded and generated futures, comparing within a context and then averaging over contexts.

Two consequences follow, one a strength and one a limitation.
First, any error that depends only on the context cancels.
A quantity that is fixed by the context has zero covariance with the centered perturbation feature, so it contributes to neither side of \eqref{eq:icm-rationale-covariance-error}.
For example, if the model renders every window of some scene slightly too dark, that offset does not enter the test (Proposition~1 in Appendix~\ref{sec:icm-analysis-centering}).
Second, the test is only as informative as its features.
A video feature that mostly tracks the arm will mostly supervise arm motion rather than object motion, and a response that no feature in the bank detects is not constrained at all.
The covariances are also estimated from finitely many windows, so finite-sample noise remains.

\subsection{A worked example with linear responses}
\label{sec:icm-rationale-response}

To see what the loss enforces, we take the simplest setting in which a response to the perturbation can be written down.
Let the perturbation be a scalar $u_t\in\mathbb{R}$ with $\mathbb{E}[u_t\mid\mathcal{H}_t]=0$ and $\mathbb{E}[u_t^2\mid\mathcal{H}_t]=\sigma^2>0$, and suppose the recorded and generated features respond linearly to it,
\begin{align}
  f(\bm{O}_t) &= \alpha(\mathcal{H}_t)+\beta u_t+\xi, &
  \mathbb{E}[u_t\xi\mid\mathcal{H}_t]&=0,
  \label{eq:icm-rationale-model-rec}\\
  f(\hat{\bm{O}}_t) &= \hat\alpha(\mathcal{H}_t)+\hat\beta u_t+\hat\xi, &
  \mathbb{E}[u_t\hat\xi\mid\mathcal{H}_t]&=0.
  \label{eq:icm-rationale-model-gen}
\end{align}
Here $\alpha$ and $\hat\alpha$ absorb everything the context determines, the response coefficients $\beta$ and $\hat\beta$ say how far the feature moves per unit of perturbation, and the noise terms $\xi$ and $\hat\xi$ are uncorrelated with the perturbation.
A model that responds to actions as the real scene does has $\hat\beta=\beta$.

\paragraph{The signed test.}
Take the signed test $g=u_t$, and assume for now that $\beta$, $\hat\beta$ and $\sigma^2$ are constants.
Substituting the two responses and taking the expectation within a context gives three terms,
\begin{equation}
  \mathbb{E}[u_t\{f(\bm{O}_t)-f(\hat{\bm{O}}_t)\}\mid\mathcal{H}_t]
  = (\alpha-\hat\alpha)\underbrace{\mathbb{E}[u_t\mid\mathcal{H}_t]}_{=\,0}
    +(\beta-\hat\beta)\underbrace{\mathbb{E}[u_t^2\mid\mathcal{H}_t]}_{=\,\sigma^2}
    +\underbrace{\mathbb{E}[u_t(\xi-\hat\xi)\mid\mathcal{H}_t]}_{=\,0}.
  \label{eq:icm-rationale-three-terms}
\end{equation}
The context terms drop out because $u_t$ is centered, and the noise terms drop out because they are uncorrelated with $u_t$.
Only the gap between the two response coefficients remains, so that
\begin{equation}
  \mathbb{E}[u_t\{f(\bm{O}_t)-f(\hat{\bm{O}}_t)\}]
    = \sigma^2(\beta-\hat\beta),
  \qquad
  \left(\mathbb{E}[u_t\{f(\bm{O}_t)-f(\hat{\bm{O}}_t)\}]\right)^2
    = \sigma^4(\beta-\hat\beta)^2.
  \label{eq:icm-rationale-linear}
\end{equation}
The squared term, which is this test's contribution to $\mathcal{L}_{\mathrm{AFFECT}}$ up to its scale, is zero exactly when $\hat\beta=\beta$.

When $\beta\neq0$, the three typical failures are all penalized.
A model that ignores the action ($\hat\beta=0$) and a model that doubles the response ($\hat\beta=2\beta$) both incur $\sigma^4\beta^2$, and a model that responds in the wrong direction ($\hat\beta=-\beta$) incurs four times as much.
In each case the penalty is proportional to the squared distance of $\hat\beta$ from $\beta$.
The target is the recorded response itself, not a larger one.
Objectives that reward separating futures under different actions behave differently, since a stronger response always helps them.

\paragraph{Context-dependent responses.}
The constant coefficients were a simplification.
If the coefficients and the perturbation variance depend on the context, the same calculation gives the discrepancy $\mathbb{E}[\sigma^2(\mathcal{H}_t)\{\beta(\mathcal{H}_t)-\hat\beta(\mathcal{H}_t)\}]$.
The average over contexts is taken before squaring, so a model that overreacts in some contexts and underreacts in others can partly cancel.
The loss therefore constrains the variance-weighted average response error, not the error in every context separately.

\paragraph{Nonlinear responses and the magnitude test.}
The linear model illustrates what the statistic measures, and it is not an assumption about real dynamics.
To look one step beyond it, suppose the perturbation is symmetric about zero given $\mathcal{H}_t$, and add a quadratic term to each of the two responses,
\begin{align}
  f(\bm{O}_t) &= \alpha(\mathcal{H}_t)+\beta u_t+\gamma\,(u_t^2-\sigma^2)+\xi,
  \label{eq:icm-rationale-quad-rec}\\
  f(\hat{\bm{O}}_t) &= \hat\alpha(\mathcal{H}_t)+\hat\beta u_t+\hat\gamma\,(u_t^2-\sigma^2)+\hat\xi,
  \label{eq:icm-rationale-quad-gen}
\end{align}
with constant $\gamma$ and $\hat\gamma$, and with noise that is uncorrelated with both $u_t$ and $u_t^2-\sigma^2$.
The new coefficients describe an even response, one that depends on how large the perturbation is but not on its direction.
Symmetry makes all odd moments of $u_t$ vanish, and this separates the signed test $g=u_t$ from the magnitude test $g=u_t^2-\sigma^2$,
\begin{align}
  \mathbb{E}[u_t\{f(\bm{O}_t)-f(\hat{\bm{O}}_t)\}]
    &= \sigma^2(\beta-\hat\beta)
    &&\text{(signed test)},
  \label{eq:icm-rationale-signed}\\
  \mathbb{E}[(u_t^2-\sigma^2)\{f(\bm{O}_t)-f(\hat{\bm{O}}_t)\}]
    &= (\gamma-\hat\gamma)\,\mathbb{E}[\operatorname{Var}(u_t^2\mid\mathcal{H}_t)]
    &&\text{(magnitude test)}.
  \label{eq:icm-rationale-magnitude}
\end{align}
The signed test ignores the quadratic terms, since $\mathbb{E}[u_t(u_t^2-\sigma^2)\mid\mathcal{H}_t]=\mathbb{E}[u_t^3\mid\mathcal{H}_t]-\sigma^2\,\mathbb{E}[u_t\mid\mathcal{H}_t]=0$.
The same moment appears when the magnitude test multiplies the linear terms, so the magnitude test ignores those and retains only $\mathbb{E}[(u_t^2-\sigma^2)^2\mid\mathcal{H}_t]=\operatorname{Var}(u_t^2\mid\mathcal{H}_t)$.
The two tests thus measure different parts of the response.
A purely even response, such as an object that is knocked over by a deviation in either direction, is invisible to the signed test, which is why the bank also contains magnitude features (Appendix~\ref{sec:icm-bank}).

\subsection{Formal results}
\label{sec:icm-analysis}

This subsection states the results behind the discussion above, together with their assumptions.
Proposition~1 makes precise the claim that centering removes context-only terms.
Proposition~2 shows that the loss is a squared MMD and that it bounds every linear combination of the tests.
Both results concern the population objective, meaning expectations rather than minibatch estimates, and say nothing about whether neural-network training reaches its minimum.

We make the following standing assumptions.
\begin{itemize}[leftmargin=*,itemsep=1pt,topsep=3pt]
\item The feature maps and the positive scales $s_b$ are fixed.
Since the scales are estimated from recorded windows before \method{} training, the results strictly hold conditional on those estimates, or exactly if the scales come from a separate calibration set.
\item All displayed expectations are finite, and every feature is square-integrable wherever we use a variance or an $L^2$ projection.
\end{itemize}
The conditioning set $\mathcal{H}_t$ is as defined in Appendix~\ref{sec:icm-rationale-moments}.
Following Appendix~\ref{sec:icm-objective}, we index tests by $b\in\mathcal{B}$ with $|\mathcal{B}|=K_\psi K_\phi$, where each $b$ corresponds to one pair $(i,j)$ of the main text and consists of a centered perturbation feature $g_b$ and a video feature $f_b$.

\subsubsection{Centering removes context-only terms}
\label{sec:icm-analysis-centering}

Section~\ref{sec:method-statistics} claims that centering the perturbation features makes static appearance drop out of the loss.
Proposition~1 makes this precise.
It shows that a centered feature is uncorrelated with everything the context determines, and that centering removes exactly the part of a feature that the context can predict.

\paragraph{Proposition 1 (centering removes context-only components).}
Let $w=w(\bm{U}_t;\mathcal{H}_t)$ be a square-integrable perturbation feature and let $g=w-\mathbb{E}[w\mid\mathcal{H}_t]$ be its centered version.
\begin{enumerate}[label=(\roman*),leftmargin=*,itemsep=1pt,topsep=3pt]
\item The centered feature is orthogonal in $L^2$ to every square-integrable function of the context, that is, $\mathbb{E}[g\,\alpha(\mathcal{H}_t)]=0$ for every such function $\alpha$.
\item The conditional mean $\mathbb{E}[w\mid\mathcal{H}_t]$ is the unique (up to almost-sure equality) minimizer of $\mathbb{E}[(w-\alpha(\mathcal{H}_t))^2]$ over square-integrable functions $\alpha$ of $\mathcal{H}_t$.
\end{enumerate}

\paragraph{Interpretation.}
Part (i) is the property that the loss relies on.
Whatever the context determines about a video feature has zero expected product with $g$, whether it appears in the recorded or in the generated future.
Part (ii) says that the conditional mean is the best prediction of $w$ from the context.
Subtracting it therefore removes everything about the feature that is predictable before the window, and leaves only its variation within a context.

\paragraph{Proof.}
Write $\bar w=\mathbb{E}[w\mid\mathcal{H}_t]$, so that $g=w-\bar w$.
By conditional Jensen, $\mathbb{E}[\bar w^2]\leq\mathbb{E}[w^2]<\infty$, so $\bar w$ and hence $g$ are square-integrable.
Since $\bar w$ is a function of $\mathcal{H}_t$, it can be taken out of the conditional expectation, which gives
\begin{equation}
  \mathbb{E}[g\mid\mathcal{H}_t]
  =\mathbb{E}[w\mid\mathcal{H}_t]-\bar w
  =0.
\end{equation}
For part (i), let $\alpha=\alpha(\mathcal{H}_t)$ be square-integrable.
The product $g\alpha$ is integrable by Cauchy--Schwarz, so the tower property applies.
Because $\alpha$ is also a function of $\mathcal{H}_t$, it too can be taken out of the inner expectation, which gives
\begin{equation}
  \mathbb{E}[g\alpha]
  =\mathbb{E}\bigl[\mathbb{E}[g\alpha\mid\mathcal{H}_t]\bigr]
  =\mathbb{E}\bigl[\alpha\,\mathbb{E}[g\mid\mathcal{H}_t]\bigr]
  =\mathbb{E}[\alpha\cdot 0]
  =0.
\end{equation}

For part (ii), fix a square-integrable $\alpha(\mathcal{H}_t)$ and split the residual as $w-\alpha=g+(\bar w-\alpha)$.
Expanding the square gives
\begin{equation}
  \mathbb{E}[(w-\alpha)^2]
  =\mathbb{E}[g^2]
   +2\,\mathbb{E}[g\,(\bar w-\alpha)]
   +\mathbb{E}[(\bar w-\alpha)^2].
\end{equation}
The function $\bar w-\alpha$ depends only on $\mathcal{H}_t$ and is square-integrable, so part (i) makes the cross term zero.
Hence
\begin{equation}
  \mathbb{E}[(w-\alpha)^2]
  =\mathbb{E}[g^2]+\mathbb{E}[(\bar w-\alpha)^2]
  \;\geq\;\mathbb{E}[g^2]
  =\mathbb{E}[(w-\bar w)^2].
\end{equation}
The first term does not depend on $\alpha$, so the minimum is reached at $\alpha=\bar w$.
Equality holds if and only if $\mathbb{E}[(\bar w-\alpha)^2]=0$, that is, if and only if $\alpha=\bar w$ almost surely, which gives uniqueness. \hfill$\square$

\paragraph{Consequence for the loss.}
We now apply part (i) to a test with centered perturbation feature $g$ and video feature $f$.
Suppose the generated feature contains a context-only error, so that $f(\hat{\bm{O}}_t)=f(\bm{O}_t)+r(\mathcal{H}_t)$ with $r$ square-integrable, as when the model renders a scene $\bm{o}_t$ slightly too dark.
Then
\begin{equation}
  \mathbb{E}\bigl[g\{f(\bm{O}_t)-f(\hat{\bm{O}}_t)\}\bigr]
  =-\mathbb{E}[g\,r(\mathcal{H}_t)]
  =0,
\end{equation}
so the error contributes nothing to the test.
By linearity the same holds for a context-only term added to any feature, recorded or generated.
This is the sense in which the loss measures what the action changed and not what the scene looks like.
It also means that \method{} cannot correct context-only errors, which remain the responsibility of the flow loss.

\subsubsection{Why the loss uses products}
\label{sec:icm-analysis-products}

Section~\ref{sec:method-statistics} rewrites the loss as a finite-feature MMD.
Proposition~2 states this formally and spells out what it buys.
A small loss bounds not only each individual test but every normalized linear combination of the tests.

\paragraph{Proposition 2 (the loss bounds all combinations of tests).}
Let $P$ denote the joint law of $(\mathcal{H}_t,\bm{U}_t,\bm{O}_t)$ in the data, and let $Q_\theta$ denote the same law with $\bm{O}_t$ replaced by the generated $\hat{\bm{O}}_t$.
Define the feature map $\Phi$ with entries $\Phi_b(\mathcal{H}_t,\bm{U}_t,\bm{O}_t)=g_b(\bm{U}_t;\mathcal{H}_t)\,f_b(\bm{O}_t)/s_b$, and let $\bm{\mu}=\mathbb{E}_P[\Phi]-\mathbb{E}_{Q_\theta}[\Phi]$.
\begin{enumerate}[label=(\roman*),leftmargin=*,itemsep=1pt,topsep=3pt]
\item The loss is the squared norm of $\bm{\mu}$, which is also the largest squared gap over unit-norm combinations of the tests,
\begin{equation}
  \mathcal{L}_{\mathrm{AFFECT}}=\|\bm{\mu}\|_2^2
  =\sup_{\|\bm{\omega}\|_2\leq1}
     \left(\mathbb{E}_P[\bm{\omega}^\top\Phi]
                -\mathbb{E}_{Q_\theta}[\bm{\omega}^\top\Phi]\right)^2.
  \label{eq:icm-proof-dual}
\end{equation}
\item Every linear combination of the tests satisfies $|\bm{\omega}^\top\bm{\mu}|\leq\|\bm{\omega}\|_2\sqrt{\mathcal{L}_{\mathrm{AFFECT}}}$.
\item The loss is the squared MMD between $P$ and $Q_\theta$ under the linear kernel $k(\zeta,\zeta')=\Phi(\zeta)^\top\Phi(\zeta')$.
\end{enumerate}

\paragraph{Interpretation.}
The vector $\bm{\mu}$ collects the scaled gaps between the recorded and the generated action effects, one per test, and the loss is its squared length.
Part (i) says that the loss equals the largest squared gap that any unit-norm combination of the tests can reveal, so no such combination can hide a gap larger than $\sqrt{\mathcal{L}_{\mathrm{AFFECT}}}$.
Part (ii) extends the guarantee to tests that are not in the bank themselves.
For example, let $g'=\sum_i a_i\,g_{t,i}$ and $f'=\sum_j c_j\,\phi_j$ be a perturbation feature and a video feature built as linear combinations of the bank.
The gap between their recorded and generated action effects is $\bm{\omega}^\top\bm{\mu}$ with $\omega_{ij}=a_i c_j s_{ij}$, so it is at most $\|\bm{\omega}\|_2\sqrt{\mathcal{L}_{\mathrm{AFFECT}}}$ in absolute value.
Part (iii) is what connects the minibatch estimator of Appendix~\ref{sec:icm-estimator} to the standard unbiased MMD estimator.

\paragraph{Proof.}
For part (i), we first show that $\mathcal{L}_{\mathrm{AFFECT}}=\|\bm{\mu}\|_2^2$.
The laws $P$ and $Q_\theta$ share the same $(\mathcal{H}_t,\bm{U}_t)$ and differ only in the future.
The $b$-th entry of $\bm{\mu}$ is therefore
\begin{equation}
  \mu_b
  =\frac{\mathbb{E}[g_b f_b(\bm{O}_t)]-\mathbb{E}[g_b f_b(\hat{\bm{O}}_t)]}{s_b}
  =\mathbb{E}\!\left[\frac{g_b}{s_b}\bigl\{f_b(\bm{O}_t)-f_b(\hat{\bm{O}}_t)\bigr\}\right]
  =\mathbb{E}[d_b],
\end{equation}
with $d_b$ as in \eqref{eq:icm-total}.
Summing the squares gives $\|\bm{\mu}\|_2^2=\sum_{b\in\mathcal{B}}(\mathbb{E}[d_b])^2=\mathcal{L}_{\mathrm{AFFECT}}$.

Next, fix $\bm{\omega}\in\mathbb{R}^{|\mathcal{B}|}$.
By linearity of expectation,
\begin{equation}
  \mathbb{E}_P[\bm{\omega}^\top\Phi]-\mathbb{E}_{Q_\theta}[\bm{\omega}^\top\Phi]
  =\sum_b \omega_b\bigl(\mathbb{E}_P[\Phi_b]-\mathbb{E}_{Q_\theta}[\Phi_b]\bigr)
  =\bm{\omega}^\top\bm{\mu}.
\end{equation}
The supremum in \eqref{eq:icm-proof-dual} is thus $\sup_{\|\bm{\omega}\|_2\leq1}(\bm{\omega}^\top\bm{\mu})^2$.
Cauchy--Schwarz gives the upper bound $(\bm{\omega}^\top\bm{\mu})^2\leq\|\bm{\omega}\|_2^2\|\bm{\mu}\|_2^2\leq\|\bm{\mu}\|_2^2$ whenever $\|\bm{\omega}\|_2\leq1$.
The bound is attained.
If $\bm{\mu}\neq\bm{0}$, then $\bm{\omega}^\star=\bm{\mu}/\|\bm{\mu}\|_2$ has unit norm and $(\bm{\omega}^{\star\top}\bm{\mu})^2=\|\bm{\mu}\|_2^2$.
If $\bm{\mu}=\bm{0}$, both sides are zero.
This proves \eqref{eq:icm-proof-dual}.

For part (ii), the case $\bm{\omega}=\bm{0}$ is trivial, and otherwise we apply the bound of part (i) to the unit vector $\bm{\omega}/\|\bm{\omega}\|_2$ and multiply by $\|\bm{\omega}\|_2$.
This gives $|\bm{\omega}^\top\bm{\mu}|\leq\|\bm{\omega}\|_2\|\bm{\mu}\|_2=\|\bm{\omega}\|_2\sqrt{\mathcal{L}_{\mathrm{AFFECT}}}$.

For part (iii), let $\zeta,\zeta'\sim P$ and $\hat\zeta,\hat\zeta'\sim Q_\theta$ be independent.
Under the linear kernel $k(\zeta,\zeta')=\Phi(\zeta)^\top\Phi(\zeta')$, the squared MMD \citep{gretton2012mmd} factorizes as
\begin{equation}
  \mathbb{E}[k(\zeta,\zeta')]-2\,\mathbb{E}[k(\zeta,\hat\zeta)]+\mathbb{E}[k(\hat\zeta,\hat\zeta')]
  =\bigl\|\mathbb{E}_P[\Phi]\bigr\|_2^2
   -2\,\mathbb{E}_P[\Phi]^\top\mathbb{E}_{Q_\theta}[\Phi]
   +\bigl\|\mathbb{E}_{Q_\theta}[\Phi]\bigr\|_2^2
  =\|\bm{\mu}\|_2^2,
\end{equation}
where the first equality uses independence to factor each expectation of an inner product. \hfill$\square$

\paragraph{Why products rather than concatenation.}
Before per-test scaling, $\Phi$ is the outer product of the perturbation and video feature vectors, so each entry tests an interaction between the two.
A simpler alternative would be to match the means of the concatenation $[\psi(\bm{U}_t),\phi(\bm{O}_t)]$, but that only compares the marginal distributions of perturbations and videos separately.
A small example, the mirrored response of Section~\ref{sec:method-statistics} in its simplest form, shows that this is not enough.
Let the scalar $u_t$ be uniform on $\{-1,1\}$, and suppose the scalar recorded future is $O_t=u_t$ while the model generates $\hat{O}_t=-u_t$.
Both $O_t$ and $\hat{O}_t$ are uniform on $\{-1,1\}$, so every additive test $h_1(u_t)+h_2(O_t)$ has the same expectation under both.
The single product test, however, gives $\mathbb{E}[u_tO_t]=1$ but $\mathbb{E}[u_t\hat{O}_t]=-1$, and so detects the reversal.

\section{Rollout visualizations}
\label{sec:rollout-viz}
\begin{figure*}[ht]
\centering\small
\includegraphics[width=\linewidth]{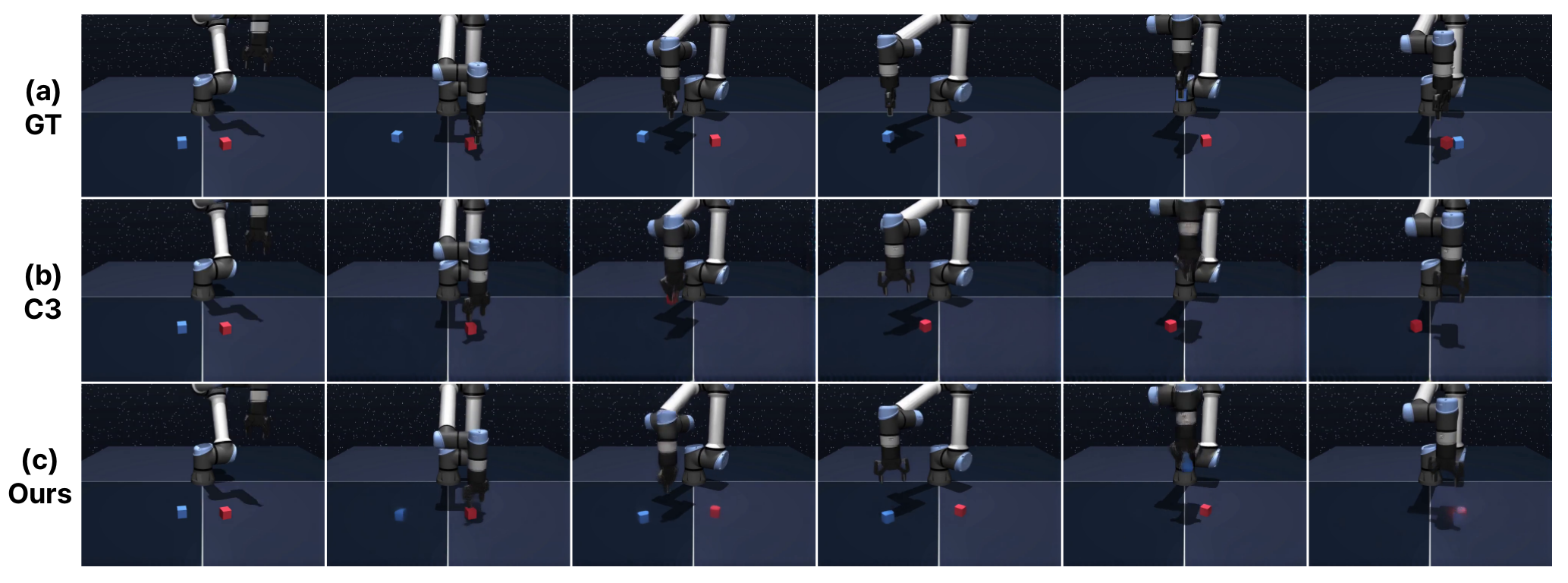}
\caption{
{\bf Open-loop long-horizon rollouts on UR5e in simulation.}
(a) Ground-truth rollout obtained by executing the action trajectory in the OGBench simulator.
(b, c) Videos generated by Cosmos3 and Cosmos3 + \method{} when conditioned on the same action trajectory.
Frames are ordered in time from left to right.
}
\label{fig:rollout-ur5e}
\end{figure*}

\begin{figure*}[ht]
\centering\small
\includegraphics[width=\linewidth]{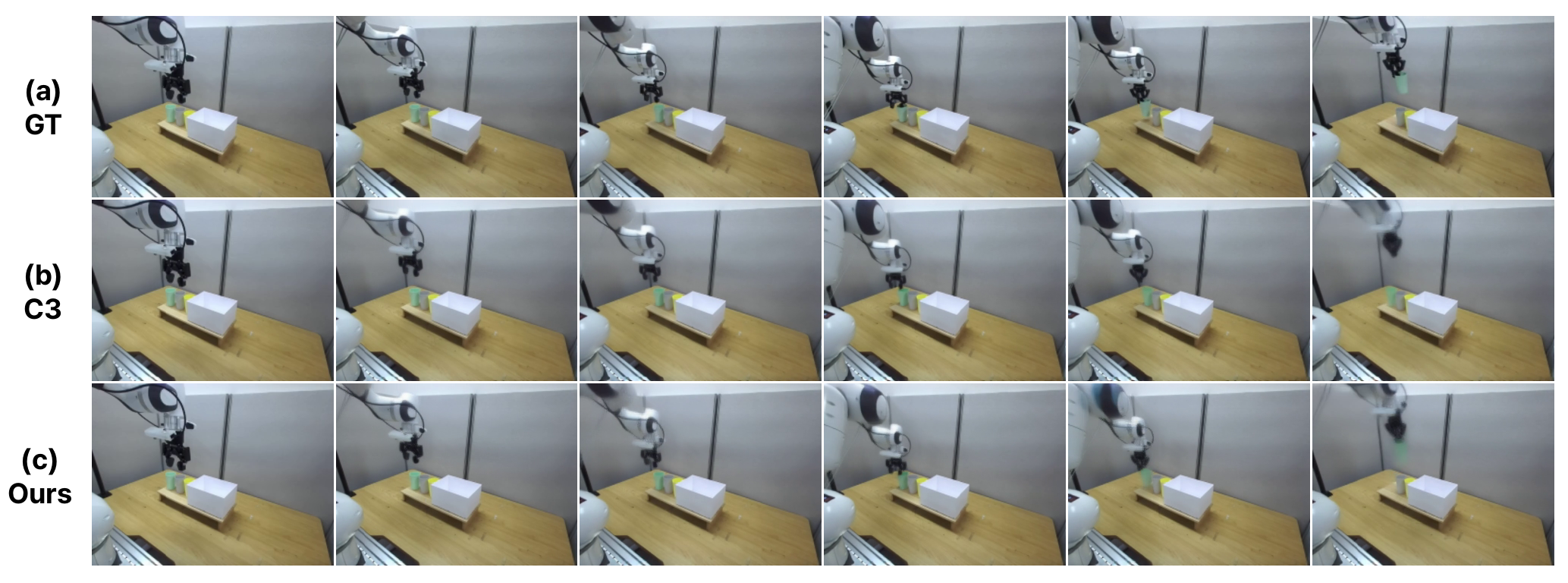}\\[6pt]
\includegraphics[width=\linewidth]{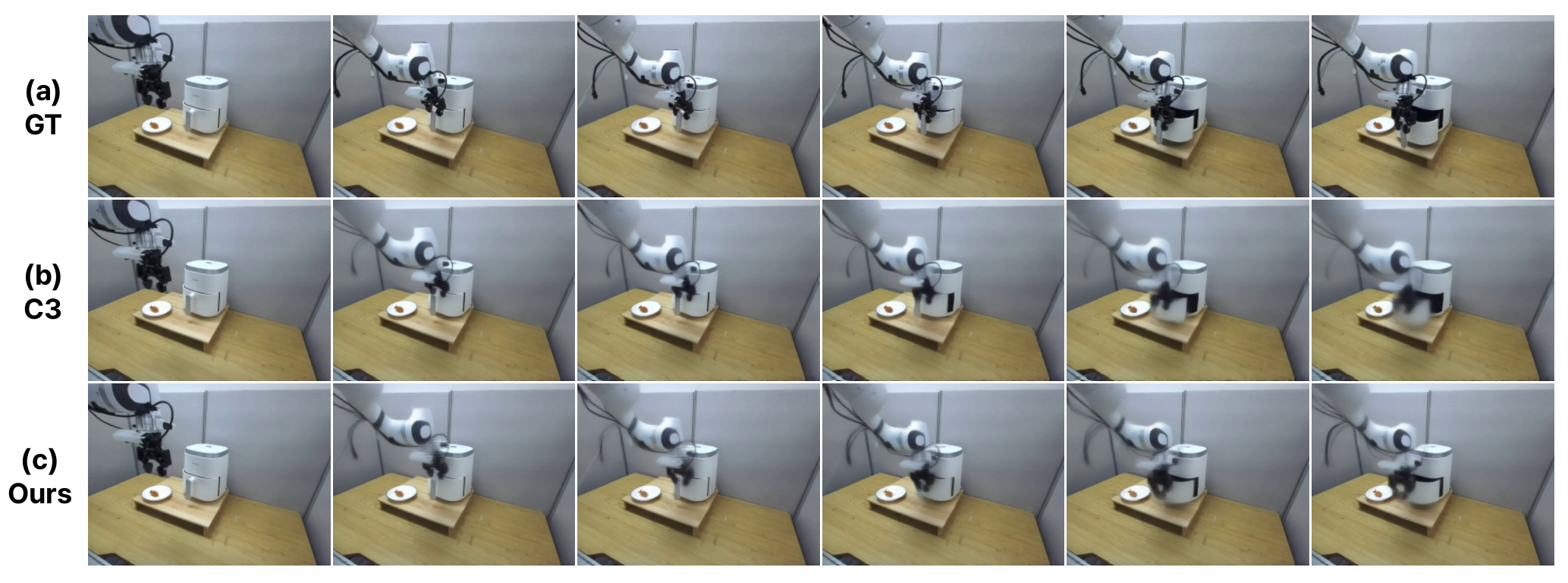}
\caption{
{\bf Open-loop long-horizon rollouts on the real-world FR3.}
Each panel shows one episode.
(a) Ground-truth rollout recorded while the FR3 executes the action trajectory.
(b, c) Videos generated by Cosmos3 and Cosmos3 + \method{} when conditioned on the same action trajectory.
Frames are ordered in time from left to right.
}
\label{fig:rollout-fr3}
\end{figure*}

We provide qualitative examples of open-loop long-horizon generation in simulation (\Figref{fig:rollout-ur5e}) and on the real-world FR3 (\Figref{fig:rollout-fr3}).
In each example, the world models start from the same initial observation as the ground-truth rollout and are conditioned on the same action trajectory, so the generated videos can be compared frame by frame against the outcome that the actions actually produced.
Since the generation is open-loop, the models receive no ground-truth frames beyond the initial observation, and any deviation from the ground-truth rollout therefore reflects how faithfully each model follows the conditioning actions over a long horizon. The results showcase that \method{} allows for better tracking of objects and results in higher quality representation of action effects.

\end{document}